\documentclass[letterpaper, 10 pt, conference]{ieeeconf}  

\IEEEoverridecommandlockouts                              

\usepackage{graphicx} 
\usepackage{booktabs}
\usepackage{array}
\usepackage{subcaption}
\makeatletter
\let\labelindent\@undefined
\makeatother
\usepackage{enumitem}
\usepackage{xcolor}
\usepackage{bm}
\usepackage{multirow}
\usepackage{url}
\usepackage{times} 
\usepackage{amsmath} 
\usepackage{amssymb}  

\usepackage{amsthm}

\usepackage{ bbold }
\usepackage{algorithm}
\usepackage[noend]{algpseudocode} 
\usepackage{hyperref}

\newtheorem{thm}{Theorem}
\newtheorem{definition}[thm]{Definition}

\title{\LARGE \bf
Multi-Robot Coordination for Planning under Context Uncertainty
}

\author{Pulkit Rustagi$^{1}$, Kyle Hollins Wray$^{2}$, and Sandhya Saisubramanian$^{1}$
\thanks{$^{1}$Collaborative Robotics and Intelligent Systems
(CoRIS) Institute, Oregon State University, Corvallis, OR 97331, USA
        {\tt\small \{rustagip, sandhya.sai\}@oregonstate.edu}}%
\thanks{$^{2}$Khoury College of Computer Sciences, Northeastern University,
        Boston, MA 02115, USA
        {\tt\small k.wray@northeastern.edu}}%
}

\begin{document}

\maketitle
\thispagestyle{empty}
\pagestyle{empty}

\begin{abstract}

Real-world robots often operate in settings where objective priorities depend on the \emph{underlying context} of operation. When the underlying context is unknown apriori, multiple robots may have to coordinate to gather informative observations to infer the context, since acting based on an incorrect context can lead to misaligned and unsafe behavior. Once the underlying true context is inferred, the robots optimize their task-specific objectives in the preference order induced by the context. We formalize this problem as a \emph{Multi-Robot Context-Uncertain Stochastic Shortest Path} (MR-CUSSP), which captures context-relevant information at landmark states through joint observations. Our two-stage solution approach is composed of: (1) \emph{CIMOP} (Coordinated Inference for Multi-Objective Planning) to compute plans that guide robots toward informative landmarks to efficiently infer the true context, and (2) \emph{LCBS} (Lexicographic Conflict-Based Search) for collision-free multi-robot path planning with lexicographic objective preferences, induced by the context. We evaluate the algorithms using three simulated domains and demonstrate its practical applicability using five mobile robots in the salp domain setup.

\end{abstract}

\section{INTRODUCTION}

Multi-robot systems in the real world must often perform multi-objective planning for task completion and robot coordination. The preference ordering over objectives is often determined by the \emph{underlying context} of operation~\cite{AWKjais24,jiang2025icra,rustagi2025multi}, defined based on factors such as resource availability, geographic, or temporal aspects of the environment. For example, salp-inspired underwater robots~\cite{yang2025effect} must prioritize minimizing ecological disturbance in coral zones over energy and speed. In areas of strong eddy currents, they must prioritize stability and energy conservation over speed. 

\begin{figure}[t]
    \centering
    \includegraphics[width=0.9\linewidth]{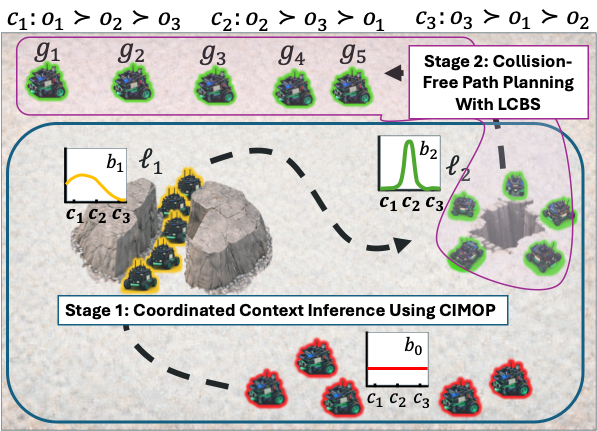} 
    \caption{Illustration of multiple GTernal robots~\cite{wilson2020robotarium} with a shared belief over true context. Accurate context-relevant observations are available only when robots are in a required configuration at a landmark (e.g., chain at $\ell_1$ and ring at $\ell_2$). After context inference, robots compute plans aligned with the context-induced objective preferences to reach their goal.}
    \label{fig:overview}
\end{figure}

When robots do not have prior knowledge of the exact underlying context, they must actively gather information before task execution, as
operating under an incorrect context can lead to inefficient coordination or unsafe behavior~\cite{merlin2024robot,merlin2025least,atanasov2015decentralized,schlotfeldt2018anytime}. Crucially, in many settings such as salps and disaster rescue~\cite{AWKjais24}, informative observations require \emph{joint sensing} in certain configurations. For example, distinguishing between a coral zone and an eddy current zone requires multiple robots to form a ring around the crevice and perform synchronized measurements of flow and particulates. Individual measurements can only measure local velocity and particulate concentration and are insufficient to infer global circulation patterns. Moreover, robots may need to repeatedly form specific configurations (e.g., ring, chain, star) to obtain 
context-revealing observations (Fig.~\ref{fig:overview}). 
This motivates our key question: \emph{how to effectively coordinate multiple robots to infer the underlying context to compute collision-free paths that optimize context-dependent objective preferences}?

Existing approaches to preference-based planning typically assume that the relevant preference ordering is known a priori, and do not model its dependence on latent context that must be inferred from observations~\cite{sharma2022correcting}. A common alternative is to compute a Pareto frontier over the objectives when their preference ordering is unknown, as widely used in multi-objective multi-agent path finding (MO-MAPF) works~\cite{wang2024efficient,ren2021multi,ren2023binary}. However, selecting a solution from the Pareto frontier for execution ultimately requires knowledge of the true context. Meanwhile, planning approaches that incorporate active information gathering for planning typically assume that robots can obtain informative observations independently~\cite{atanasov2015decentralized,schlotfeldt2018anytime}. These methods, therefore, do not support scenarios that require coordination for joint sensing by multiple robots to collect informative observations about the context. Finally, once the context and corresponding objective ordering are known, Pareto-based approaches become computationally inefficient as they fail to exploit the available preference information during planning.

We present \emph{Multi-Robot Context-Uncertain Stochastic Shortest Path} (MR-CUSSP), a framework designed to model settings where objective preferences depend on a fixed but unknown context that can only be inferred via joint observations. The robots maintain a \emph{shared} belief over possible contexts, which is updated based on observations. We consider contexts to map to fixed lexicographic preferences over objectives, and assume the existence of belief-collapsing observations at \emph{landmark states} that uniquely identify the governing preference ordering that must be followed for task completion. These belief-collapsing observations depend on the joint state and actions, enforcing robot coordination for context inference. We focus on navigation tasks in which robots must reach goal locations, while avoiding collisions with other robots and optimizing context-induced lexicographic ordering over objectives.

Our two-stage solution approach (Fig.~\ref{fig:overview}) involves computing: (1) joint plans to visit landmark states such that the underlying context can be quickly inferred; and (2) collision-free individual paths for task completion using the lexicographic ordering over objectives, induced by the inferred context. For the first stage, we present \emph{CIMOP} (Coordinated Inference for Multi-Objective Planning) which computes joint plans to visit landmark states in an order that accelerates belief collapse through informative observations. 
For the second stage, we present \emph{Lexicographic Conflict-Based Search} (LCBS), an algorithm that builds on conflict-based search (CBS)~\cite{sharon2012conflict} to compute a collision-free path for each robot, aligned with the context-induced objective ordering. LCBS uses lexicographic $A^*$ and constraint branching~\cite{wang2024efficient} to produce collision-free paths for the robots.

Our proposed two-stage structure separates the challenges of information gathering and task execution, enabling each to be solved with algorithms tailored to their structure. By first inferring the context and then planning with a known preference ordering, the approach enables scalable, collision-free robot planning. 
Empirical evaluations using three domains in simulation and five mobile robots show that our proposed two-stage approach outperforms state-of-the-art baselines in each stage.

\section{BACKGROUND AND RELATED WORKS}
\noindent\textbf{Stochastic Shortest Path (SSP) problem~}
SSPs are a popular framework for modeling goal-oriented tasks that require sequential decision making under stochastic outcomes~\cite{SWPZiros19}, such as autonomous navigation~\cite{stracca2026risk} and warehouse operation~\cite{street2025planning}. Formally, an SSP is represented by the tuple $\langle S, A, T, C, s_0, s_g\rangle$, with a finite set of states $S$, actions $A$, transition function $T$ that represents probability of reaching a successor state, a cost function $C$, and start and goal states denoted by $s_0$ and $s_g$, respectively. This work extends SSPs to model settings with multiple robots and lexicographic ordering over objectives induced by a latent context. 

\vspace{2pt}
\noindent\textbf{Active information gathering~}
Existing works on information gathering reduces uncertainty by planning informative actions~\cite{schlotfeldt2018anytime} or sampling exploration policies~\cite{kantaros2021sampling}, but assume that observations can be obtained independently by robots. A recent approach models active sensing for a single robot in partially observable settings using Locally Observable Markov Decision Processes (LOMDPs)~\cite{merlin2024robot,merlin2025least}. These approaches, however, do not model settings where observations depend on coordinated robot configurations, which is precisely the setting we target.

\vspace{2pt}
\noindent\textbf{Multi-Objective MAPF}
Multi-objective multi-agent path finding (MO-MAPF) extends multi-agent path finding (MAPF) to settings with multiple objectives, each associated with a cost function~\cite{ren2021multi,ren2022conflict}. Robots $\mathbf{R}\!=\!\{1, \dots, d\}$ move in discrete time on a graph $G = (V, E)$, with each robot $i \in \mathbf{R}$ following a path $\pi_i$ from start $s_i$ to goal $g_i$. A joint plan $\Pi\!=\!(\pi_1, \dots, \pi_d)$ is \emph{valid} if it avoids vertex and edge conflicts~\cite{stern2019multi}. Each robot incurs a cost vector $\mathbf{c}_i \in \mathbb{R}_+^n$, yielding total cost $\mathbf{C}(\Pi) = \sum_{i \in \mathbf{R}} \mathbf{c}_i$. 

MO-MAPF methods compute Pareto-optimal paths for each robot at the low level and resolving conflicts through a constraint tree at the high level. MO-CBS~\cite{ren2021multi} and BB-MO-CBS~\cite{ren2023binary} explicitly enumerate non-dominated joint plans, while BB-MO-CBS-$\varepsilon$ and BB-MO-CBS-$k$ control the size of the frontier through approximation or restriction~\cite{ren2023binary,wang2024efficient}. Scalarization and evolutionary approaches combine multiple objectives into a single scalar cost or fitness function before applying CBS~\cite{ho2023preference}. While effective for exploring trade-offs, these methods do not directly enforce a prescribed lexicographic ordering during search. Our approach instead integrates lexicographic comparison at both levels of CBS, ensuring that the returned solution is preference-aligned without frontier construction or scalar reduction.

\begin{figure*}[t]
    \centering
    \includegraphics[width=0.95\linewidth]{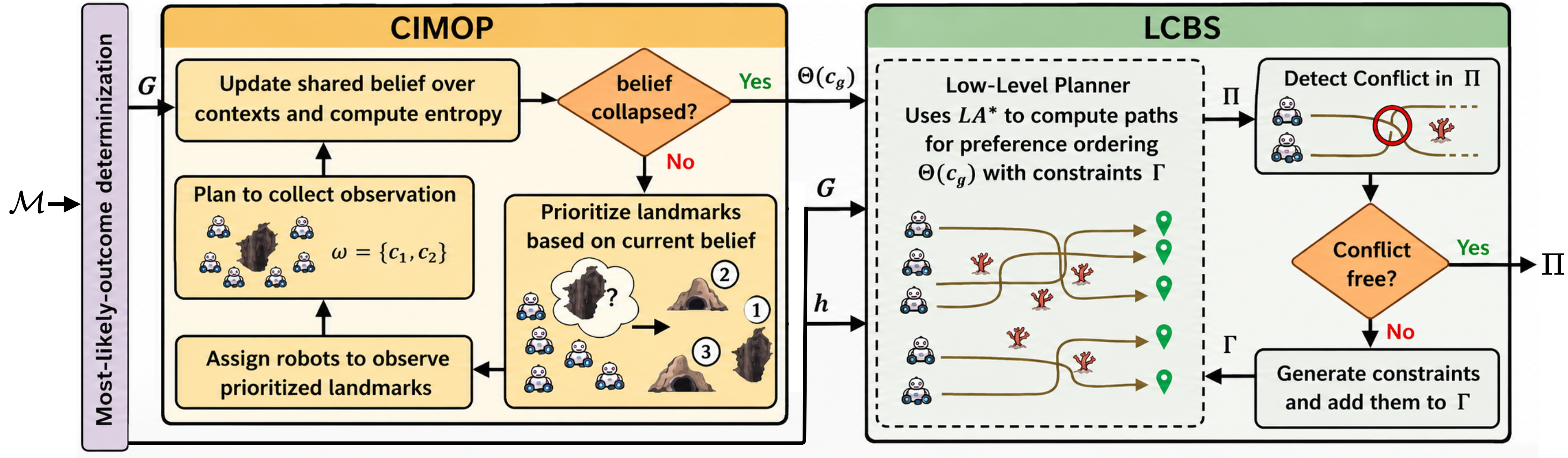}
    \caption{Solution approach overview. A most-likely-outcome determinization is first applied to obtain a discrete graph representation of the stochastic domain, enabling the use of graph-based planning methods. CIMOP prioritizes visiting landmark states that minimize the belief entropy and assigns robots accordingly, based on the current shared belief which is updated based on joint observations at landmark states. Once the context is inferred ($c_g$), the induced lexicographic ordering $\Theta(c_g)$ and a discrete graph representation of the environment, along with a heuristic, are used for task planning. LCBS uses lexicographic $A^*$ to compute preference-aligned paths, detects conflicts in the joint plan, and iteratively adds constraints using binary branching~\cite{ren2023binary} until a conflict-free solution is obtained.}
    \label{fig:solution_overview}
\end{figure*}

\section{PROBLEM FORMULATION} 
Consider a setting with $d$ robots, each with its own assigned task. The robots operate in an environment with a fixed context that is unknown a priori and must be inferred from observations obtained during execution. Since the underlying context induces a preference ordering over multiple objectives, the robots cannot successfully complete their task in a preference-aligned manner until the context is inferred. We consider tasks characterized by specific start and goal locations for each robot. The robots can independently complete their tasks but must coordinate for context inference and to avoid collisions. We formalize this setting as a \emph{Multi-Robot Context-Uncertain Stochastic Shortest Path} (MR-CUSSP) problem.

\begin{definition}
\label{def:MR-CUSSP}
A Multi-Robot Context-Uncertain Stochastic Shortest Path (MR-CUSSP) problem for a set of robots $\textbf{R}\!=\!\{1,\dots,d\}$ is defined by $\mathcal{M}\!=\!\langle \mathcal{P}, \;X, \;A, \;T,\;C, \;\Theta,\;O\rangle$ with

\begin{itemize}[leftmargin=12pt]
    \item $P = \{c_1,\dots,c_m\}$ is a finite set of possible contexts, with $c_g \in P$ as the true context initially unknown to the robots;
    \item $X = S \times P$ is the state space, where $S = \times_{k \in A} \hat{S}_k$ is the joint physical state and $P$ is the set of possible contexts;
    \item $A = \times_{k \in A} \hat{A}_k$ is the joint action space;
    \item $T : X \times A \times X \rightarrow [0,1]$ is the transition function;
    \item $C : X \times A \rightarrow \mathbb{R}^n$ is the cost for objectives $o = \{o_1,\!...,\!o_n\}$;
    \item $\Theta : P \rightarrow \theta$ maps each context $c_i \in P$ to a lexicographic ordering $\theta_i\!\in\!\theta$, where $\theta_i\!=\!C_{i1}\!\succ \cdots \succ\!C_{in}$ denotes a strict priority ordering over objectives, and $C_{ij}$ is the cost associated the $j^{th}$ priority objective under context \!$c_i$;
    \item $O : A \times X \times \Omega \rightarrow [0,1]$ is the joint observation function, where $O(a, x', \omega) = \Pr(\omega \mid a, x')$ and $\Omega = 2^P\setminus \emptyset$ is the set observations.
\end{itemize}
\end{definition}
Each state is represented by $x = \langle s, c\rangle$, with $s\in S$ and $c \in P$. The initial and goal states are denoted by $x_0 = \langle s_0, c\rangle$ and $x_g = \langle s_g, c\rangle$ with $s_0, s_g \in S$, respectively. MR-CUSSPs have mixed observable state components as $s$ is fully observable and the partial observability is restricted to the context.  Therefore, robots maintain a shared belief $b\!:\!X \rightarrow \Delta^{|X|-1}$ which is updated based on joint observations. For clarity, the rest of this paper considers homogeneous robots but MR-CUSSPs also support heterogeneous robots.

\noindent\textbf{Joint observations at landmark states} Every joint action produces an observation $\omega \in \Omega$. At \emph{landmark states}, $\mathcal{L} \subseteq S$, observations provide accurate information about a subset of potential contexts. For each $s\in \mathcal{L}$, let $\Omega_s$ denote the set of observations corresponding to  context information that can be inferred from that state. Each $\omega \in \Omega_s$ provides information about maximal set of contexts. Similar to locality-based observation models~\cite{merlin2024robot,merlin2025least}, landmark states correspond to informative physical states where observations become available only when the required joint robot configuration and actions are satisfied. For example, salp robots arranged in a ring around a crevice receive accurate flow information, reducing uncertainty over contexts. Thus,$\,\forall\,\!a\!\in\!A,\! \,x'\!=\!\langle s'\!,c'\rangle$:
\[
O(a,x',\omega) =
\begin{cases}
1, & \text{if }  s' \in \mathcal{L} \land \omega \in \Omega_{s'} \land c_g \in \omega,\\
0, & \text{if } s' \in \mathcal{L} \land \omega \in \Omega_{s'} \land  c_g \notin \omega,\\
\frac{1}{\vert \Omega \vert}, & \text{if } s' \notin \mathcal{L}.
\end{cases}
\]
Since $c_g$ is the true operating context, any observation either points to it, narrows the feasible set to one still containing it, or is uninformative. We assume no observation rules out $c_g$, so it always remains once the belief collapses.

\noindent\textbf{Belief update~} Belief $b(x)$ is semantically a belief over contexts since the physical state $s$ in $x\!=\!\langle s,c \rangle$ is fully observable. The updated belief for $x'\!=\!\langle s',c'\rangle$ after receiving an observation $\omega$ is calculated as:
\begin{eqnarray}
    b'(x'\vert b,a,\omega) &= \Pr(c'\vert b,a,\omega, s')\Pr(s'\vert b,a,\omega,s) \nonumber \\
    &= \Pr(c'\vert b,a,\omega, s')T(s,a,s') \nonumber \\
&= \eta O(a,x',\omega) b(c) T(s,a,s'),
\label{eqn:belief-update}
\end{eqnarray}
where $\eta = \Pr(\omega\vert b,a,s')^{-1}$ is a normalization constant and $b(c)$ is the belief over a context $c$. 
By definition, the observation function produces a belief-collapsing observation or no information at all. Therefore, the belief is either collapsed ($b'(c)= \{0,1\}$) or remains the same ($b'(c) = b(c)$). Since $|S|$ is finite, belief update following Eqn.~\ref{eqn:belief-update} results in a finite number of reachable beliefs for MR-CUSSP.

\begin{algorithm}[t]
\caption{CIMOP: Inference plan with belief sync}
\label{algo:CIMOP}
\begin{algorithmic}[1]
\Require Robots $\mathbf{R}$, contexts $P$, landmarks $\mathcal{L}$, init.$\,$belief$\,b_0$
\Ensure Joint plan $\Pi$ (or $\emptyset$), inferred context $c_g$

\State \textbf{Init:} $\mathbf{R}_{av}\gets \mathbf{R}$, $\mathcal{L}_{vis}\gets \{\}$, $Groups\gets [\,]$, $b\gets b_0$
\If{$H(b_0)=0$}
    \State \Return $(\emptyset,\arg\max_{c\in P}b_0)$ 
\EndIf
\State $N_{\mathcal{L}}[\ell]\gets |\mathbf{R}|\,\,\forall \ell \in \mathcal{L}$\Comment{\textcolor{gray}{initialize \#robots needed at $\ell$}}
\ForAll{$\ell \in \mathcal{L}$}
    \State $N_{\mathcal{L}}[\ell]\gets\min\!\{k|k\!=\!\arg\!\max_{k\leq|R|} \!H(b_0)\!-\!H(b'|b_0,\omega_{\ell}^k)\}$ 
\EndFor
\While{$H(b)>0$}
    \State $\mathcal{I}\gets \textsc{GetLandmarkVisitSequence}(N_{\mathcal{L}},\mathcal{L},b)$
    \For{$\ell\in\mathcal{I}\setminus\mathcal{L}_{vis}$} \Comment{\textcolor{gray}{highest to lowest visit priority}}
        \If{$|\mathbf{R}_{av}|\geq N_{\mathcal{L}}(\ell)$}
            \State $g\gets \textsc{AssignNearestRobots}(\mathbf{R}_{av},N_{\mathcal{L}}[\ell])$
            \State $g.\ell \gets \ell$ 
            \State $Groups.$\textsc{Append}($g$)
            \State $\mathbf{R}_{av}\gets\mathbf{R}_{av}\!\setminus\!g.\mathbf{R}$
        \EndIf
    \EndFor
    \State $\Pi.\textsc{Append}(\textsc{PlanToLandmark}(\mathbf{R},Groups))$
    \For{$g\in Groups$} 
        \If{$g$ reached $g.\ell$}
            \State $\mathbf{R}_{av}\gets\mathbf{R}_{av}\cup g$
            \State $\mathcal{L}_{vis}\gets \mathcal{L}_{vis}.\textsc{Append}(g.\ell)$
            \State $b\gets \textsc{UpdateBelief}(\Pi)$ \Comment{\textcolor{gray}{shared belief}}
        \EndIf
    \EndFor
\EndWhile

\State $c_g\gets\arg\max_{c\in P}b$
\State \Return $\Theta(c_g)$\Comment{\textcolor{gray}{preferences under $c_g$}}

\end{algorithmic}
\end{algorithm}

\noindent\textbf{Belief entropy}
Efficient task completion requires quick context inference by optimizing the visitation order of landmark states, based on current uncertainty over contexts. To quantify current uncertainty, we define \emph{belief entropy} as the number of contexts that remain feasible under belief $b$:
\begin{equation}
    \label{eq:entropy}
    H(b) \triangleq \sum_{x \in X} \mathbb{1}\!\left[ b(x) > 0\right] - 1.
\end{equation}
We use $H(b)$ to denote the entropy associated with a belief $b$ and $H(b'|b,\omega_\ell^k)$ to denote the entropy associated with an updated belief $b'$ as a result of joint observation ($\omega_\ell^k$) made by $k$ robots at landmark $\ell$. 
By Eqn.~\ref{eq:entropy}, $H(b)=0$ only when the belief is non-zero for exactly one context, which indicates the inference of the true context, as belief collapse with incorrect context is impossible under our observation function. 

In the following section, we present our two-stage solution approach that first infers the underlying context and then computes collision-free paths for task completion, using the context-induced objective ordering.

\section{SOLUTION APPROACH}
Solving MR-CUSSPs involves four high-level steps (Figure~\ref{fig:solution_overview}): (1) identify informative sequence of landmark states for fast context inference, and assign robot groups to visit them; (2) visit the assigned landmarks and update shared belief based on observations; (3) repeat steps (1)-(2) until belief collapse; and (4) once the true context is inferred, plan for task completion under induced objective preferences. 
For steps (1) and (2), we present an algorithm, \emph{CIMOP} (Coordinated Inference for Multi-Objective Planning). CIMOP (Alg.\ref{algo:CIMOP}) determines the order in which landmarks states should be visited, based on the initial belief, and computes coordinated plans to obtain informative observations at landmark states. For step (4), we present Lexicographic Conflict-Based Search (LCBS) that computes a plan for each robot independently, while avoiding collisions with other robots. To enable checking for potential collisions between robots in steps (2) and (4), we use most-likely outcome determinization~\cite{saisubramanian2019adaptive} to construct a deterministic approximation by using the most likely successor for each state-action pair, during planning. In addition to speeding up task planning, determinization also helps with collision avoidance as CBS-based MAPF solvers, including ours, require a graph representation with single action outcome. At execution time, robots may reach a state for which they do not have a prescribed action, as they operate under the true dynamics, which is addressed by replanning.

\subsection{Coordinated Context Inference using CIMOP}
Alg.~\ref{algo:CIMOP} first initializes the available robot set, visited landmark set, and active robot groups (Line~1). If $H(b_0)>0$, we compute the minimum number of robots required to obtain an informative observation at each landmark, denoted by $N_{\mathcal{L}}[\ell]$ (Lines~4--6). This is computed by searching over possible team sizes to find the smallest $k$ that yields the maximum entropy reduction $H(b_0)-H(b'|\omega_{\ell}^k)$. 

CIMOP iteratively computes a visitation sequence $\mathcal{I}$ over landmark states, based on current belief (Lines 7,8). Specifically, CIMOP uses Alg.~\ref{algo:GetLandmarkVisitSequence} to compute $\mathcal{I}$ by estimating the reduction in entropy that can be achieved with ${N_{\mathcal{L}}[\ell]}$ robots at a landmark $\ell$, $H(b) - H\!\big(b' \mid b,\ \omega_\ell^{N_{\mathcal{L}}[\ell]}\big)$. The landmarks are then sorted in the decreasing order of entropy reduction and used for robot assignment. Note that $N_{\mathcal{L}}[\ell]$ is determined using $b_0$ (Line~6, Alg.~\ref{algo:CIMOP}) while $\mathcal{I}$ is determined using $b$ (Line~1, Alg.~\ref{algo:GetLandmarkVisitSequence}). 
If $|\mathbf{R}_{av}|\geq N_{\mathcal{L}}[\ell]$, the nearest $N_{\mathcal{L}}[\ell]$ robots are assigned to unvisited $\ell \in \mathcal{I}$ (Lines 9-11). The assigned landmark is recorded within the group object and its robots are removed from the available pool (Lines~12-14). The process continues until either available robots are insufficient or all prioritized landmarks are examined. While any planner can be used to compute a joint plan for the group to reach its assigned landmark, our experiments use a standard conflict-based search (CBS) planner with determinization and replanning as needed~\cite{sharon2012conflict} (Line~15). 
When a group $g$ reaches its assigned landmark $g.\ell$, the robots are returned to the available pool (Line~18), the landmark is marked as visited (Line~19), and the shared belief is updated (Line~20). Since the belief does not change in non-landmark states, following our observation function, it is sufficient to update the shared beliefs when a landmark is visited.

The process of recomputing landmark priorities under the updated belief and reallocating robots is repeated until $H(b)=0$ (Lines~7-20). Once belief collapses, the context is inferred as $c_g=\arg\max_{c\in P} b$, and associated lexicographic ordering is returned for task planning (Lines~21-22).

\begin{algorithm}[t]
\caption{\textsc{GetLandmarkVisitSequence}}
\label{algo:GetLandmarkVisitSequence}
\begin{algorithmic}[1]
\Require landmark requirements $N_{\mathcal{L}}$, landmarks $\mathcal{L}$, belief $b$
\Ensure Visitation sequence $\mathcal{I}$
\State $\forall \ell\in\mathcal{L}: I_H(\ell) \gets H(b) - H\!\big(b' \mid b,\ \omega_\ell^{N_{\mathcal{L}}[\ell]}\big)$ 
\State $\mathcal{I}\gets \mathcal{L}.\textsc{Sort}(I_H,\text{order = descending})$ 

\State \Return $\mathcal{I}$
\end{algorithmic}
\end{algorithm}

\subsection{Multi-Robot Planning under Inferred Context using LCBS}
To plan under the inferred lexicographic preferences $\Theta(c_g)$, we present \emph{Lexicographic Conflict-Based Search} (LCBS), an algorithm that extends the two-level framework of Conflict-Based Search (CBS)~\cite{sharon2012conflict} to settings with strict priority ordering over multiple objectives. In LCBS, the low-level search computes a plan for each robot independently; the high-level search detects conflicts in the joint plan and iteratively generates constraints for the low-level planner. Any heuristic can be used in practice but optimality guarantees depend on the heuristic admissibility. In our experiments, we use Euclidean distance to the goal as the heuristic. 

\begin{algorithm}[t]
\caption{LA$^*$ (Lexicographic A$^*$)}
\label{alg:LA*}
\textbf{Input}: $G=(V,E)$, start-goal $(s,g)$, edge cost $\mathbf{c}_e:E\!\to\!\mathbb{R}_+^d$, heuristic $\mathbf{h}:V\!\to\!\mathbb{R}_+^d$, constraints $\Gamma$\\
\textbf{Output}: Optimal path $\pi$ from $s$ to $g$ (or $\emptyset$)
\begin{algorithmic}[1]

\State \textbf{Init:} timed states $z=(v,t)$ with $v\in V$, $t\in\mathbb{N}$; 
\State $z_0 \gets (s,0)$;\ $\mathbf{g}(z_0)\gets \mathbf{0}$; $\mathbf{f}(z_0)\gets \mathbf{h}(s)$;
\State  open list $\mathcal{O}\gets \{\}$; cost map $\mathcal{C}\gets\{\}$; plan $\pi\gets\{\}$;
\State $\mathcal{O}.\textsc{Push}(z_0)$; \ $\mathcal{C}[z_0]\gets \mathbf{0}$; 
\While{$\mathcal{O}\neq\emptyset$}
  \State $z=(v,t) \gets \mathcal{O}.\textsc{PopMin}()$ \Comment{\textcolor{gray}{lex-min $\mathbf{f}$}}
  \If{$v=g$ \textbf{and} \textbf{not }\textsc{Violates}$(z,\Gamma)$}
     \State $\pi\gets$ \textsc{ReconstructPath}$(z)$

     \State \textbf{break}
  \EndIf
  \For{each $u \in \textsc{Successors}(v)$}
     \State $y \gets (u,t{+}1)$
     \If{\textsc{Violates}$(z\!\to\!y,\Gamma)$} 
     \State \textbf{continue} \EndIf
     \State $\mathbf{g}'(y) \gets \mathbf{g}(z) + \mathbf{c}_e(v,u)$;\ \ $\mathbf{f}'(y) \gets \mathbf{g}'(y) + \mathbf{h}(u)$
     \If{$y \notin \mathcal{C}$ \textbf{or} $\mathbf{g}'(y) <_{\text{lex}} \mathcal{C}[y]$}
        \State $\mathcal{C}[y]\gets \mathbf{g}'(y)$; \textsc{Parent}$(y)\!\gets\!z$; \!$\mathbf{f}(y)\gets \mathbf{f}'(y)$
        \State $\mathcal{O}.\textsc{Push}(y)$
     \EndIf
  \EndFor
\EndWhile
\State \Return $\pi$
\end{algorithmic}
\end{algorithm}

\begin{algorithm}[t]
\caption{LCBS (High-Level Search)}
\label{alg:LCBS}
\textbf{Input}: $G=(V,E)$, robots $\mathbf{R}=\{1,\dots,d\}$, start-goal pairs $(s^i,g^i)\,\forall i\in \mathbf{R}$, edge cost $\mathbf{c}_e$, heuristic $\mathbf{h}$,  preferences $\Theta$\\
\textbf{Output}: Conflict-free joint plan $\Pi=(\pi_1,\dots,\pi_d)$ (or $\emptyset$)
\begin{algorithmic}[1]
\State \textbf{Init:}  $<_{\text{lex}}\gets\Theta$; $\mathcal{N}_0\gets\emptyset$; $\mathcal{O}_{\mathrm{HL}}\gets\{\}$;
\State \textbf{Init:}  $\Gamma_0\gets\{\}$;  $\pi_i\gets \emptyset \;\forall i\in\mathbf{R}$;
\For{each robot $i \in \mathbf{R}$}
   \State $\pi_i \gets \textsc{LA*}(G,s^i,g^i,\mathbf{c}_e,\mathbf{h},\Gamma_0)$
   \If{$\pi_i=\emptyset$} \State \Return $\emptyset$ \EndIf
\EndFor
\State $\mathcal{N}_0.\Pi \gets [\pi_i]_{i\in \mathbf{R}}$; $\mathcal{N}_0.\mathbf{C} \gets \sum_{i}\mathbf{c}_e(\pi_i)$; $\mathcal{O}_{\mathrm{HL}}.\textsc{Push}(\mathcal{N}_0)$;
\While{$\mathcal{O}_{\mathrm{HL}}\neq\emptyset$}
   \State $\mathcal{N} \gets \mathcal{O}_{\mathrm{HL}}.\textsc{PopMin}()$ \Comment{\textcolor{gray}{lex-min $\mathbf{C}(\mathcal{N}.\Pi)$}}
   \State $\textit{conflict} \gets \textsc{DetectFirstConflict}(\mathcal{N}.\Pi)$ 
   \If{$\textit{conflict}=\emptyset$} \State \Return$\Pi\gets\mathcal{N}.\Pi$ 
   \Comment{\textcolor{gray}{return conflict-free plan}}\EndIf 
   \State $(\textit{type},loc,time,id_i,id_j) \gets \textit{conflict}$ 
   \State $\{\gamma^i,\gamma^j\}\gets \textsc{GenerateConstraints}(\textit{conflict})$ 
   \For{each $a \in \{i,j\}$}
      \State Create child $\mathcal{N}_a$;\ \ $\mathcal{N}_a.\Gamma \gets \mathcal{N}.\Gamma \cup \{\gamma^a\}$
      \State $\pi_a' \gets \textsc{LA*}(G,s^a,g^a,\mathbf{c}_e,\mathbf{h},\mathcal{N}_a.\Gamma)$
      \If{$\pi_a' \neq \emptyset$}
          \State $\mathcal{N}_a.\Pi[a] \gets \pi_a'$; $\mathcal{N}_a.\mathbf{C} \gets \sum_{i}\mathbf{c}_e(\pi_i)$
         \State $\mathcal{O}_{\mathrm{HL}}.\textsc{Push}(\mathcal{N}_a)$
      \EndIf
   \EndFor
\EndWhile
\State \Return $\Pi$
\end{algorithmic}
\end{algorithm}

To compare two cost vectors $\mathbf{u}$ and $\mathbf{v}$, we define $\mathbf{u} <_{\text{lex}} \mathbf{v}$ as the lexicographic comparison under $\Theta$. Specifically, $\mathbf{u} <_{\text{lex}} \mathbf{v}$ iff $\exists j \in \{1,\dots,n\}$ s.t. $
\mathbf{u}^j < \mathbf{v}^j \text{ and } \forall k<j,\; \mathbf{u}^k = \mathbf{v}^k.$  
The $<_{\text{lex}}$ is used by LCBS to prune dominated cost vectors during planning. 

\vspace{2pt}
\noindent\textbf{Low-Level Search~} The low-level planner in LCBS is \emph{lexicographic A$^*$} (LA$^*$) (Alg.~\ref{alg:LA*}), which computes individual robot paths under the ordering induced by $\Theta$. 
LA$^*$ searches over time-augmented states $z=(v,t)$ with cumulative vector cost $\mathbf{g}(v,t)$ and admissible heuristic $\mathbf{h}(v)$. The evaluation key is $\mathbf{f}(v,t)=\mathbf{g}(v,t)+\mathbf{h}(v)$. The open list $\mathcal{O}$ is queried by $\textsc{PopMin}$ to extract the lexicographically smallest $\mathbf{f}$ under $<_{\text{lex}}$ (Line~6), so states with lexicographically smaller $\mathbf{f}$ are expanded first.

Alg.~\ref{alg:LA*} initializes $\mathcal{O}$ with $(s,0)$ and maintains a closed map $\mathcal{C}$ storing the best $\mathbf{g}$ per state (Lines~4). If the popped state reaches the goal and satisfies constraints $\Gamma$, the path is returned (Lines~6-9, 18). Otherwise, successor states from outgoing edges are generated, and transitions violating $\Gamma$ are skipped (Lines~10-13). A successor is inserted only if its $\mathbf{g}$ improves under $<_{\text{lex}}$ (Line~14-17), ensuring that each state maintains a single best cost vector following $\Theta$.

Optimizing a lexicographic ordering using LA* is more than just a tie-breaking. As node selection (Line~6) and state updates (Line~15) are governed by $<_{\text{lex}}$,  lexicographically-worse path candidates for a state are discarded immediately. The search, thus, directly and efficiently optimizes the strict priority ordering specified by $\Theta$.

\vspace{2pt}
\noindent\textbf{High-Level Search~} For the high-level search (Alg.~\ref{alg:LCBS}), we adopt the constraint-tree (CT) framework from BB-MO-CBS-pex~\cite{wang2024efficient}. 
Each CT node denotes a joint plan $\mathcal{N}.\Pi$, joint cost vector $\mathcal{N}.\mathbf{C}$, and constraint set $\mathcal{N}.\Gamma$. The lexicographic comparator $<_{\text{lex}}$ is first defined according to $\Theta$, along side initialization of root node $\mathcal{N}_0$, high-level open list $\mathcal{O}_{\mathrm{HL}}$, constraint set $\Gamma_0$ and robot policies $\pi_{i}\,\forall i \in \mathbf{R}$ (Line~1, 2). The high-level open list $\mathcal{O}_{\mathrm{HL}}$ is ordered lexicographically by $\mathcal{N}.\mathbf{C}$ (Line~9). The root node is constructed by invoking LA$^*$ for each robot under empty constraints (Lines~3--6), and the joint cost is computed, $\mathcal{N}_0.\mathbf{C}=\sum_{i\in \mathbf{R}}\mathbf{c}_e(\pi_i)$ (Line~7). At each iteration, the node with lexicographically smallest joint cost is popped (Line~9). If no conflict is detected, the joint plan is returned (Lines~10--12). Otherwise, the earliest conflict is identified and the node is split into two child nodes, each imposing a constraint on one of the conflicting robots (Lines~13--16). The affected robot is replanned using LA$^*$ under the updated constraint set (Line~17). Feasible children update their joint plan and cost (Lines~18--19) and are pushed into $\mathcal{O}_{\mathrm{HL}}$ (Line~20).

The first conflict-free CT node popped is lexicographically optimal as both high-level and low-level search order nodes under $<_{\text{lex}}$ and LA$^*$ returns lexicographically optimal paths under constraints. 

\newcolumntype{C}[1]{>{\centering\arraybackslash}m{#1}}
\begin{figure*}[t]
\centering
\setlength{\tabcolsep}{2pt}          
\renewcommand{\arraystretch}{1.0}     

\begin{tabular}{C{0.13\textwidth} C{0.19\textwidth} C{0.20\textwidth} C{0.22\textwidth} C{0.20\textwidth}}
\toprule
\text{{\footnotesize Map}} &
\text{{\footnotesize (i) Entropy for 5 robots ($\downarrow$)}} &
\text{{\footnotesize (ii) Entropy for 35 robots ($\downarrow$)}} &
\text{{\footnotesize (iii) Cumulative entropy for 5 robots ($\downarrow$)}} &
\text{{\footnotesize \qquad (iv) Planning Time ($\downarrow$)}} \\
\midrule

\begin{minipage}[c]{\linewidth}\centering
\vspace{-7mm}
  \includegraphics[width=0.8\linewidth]{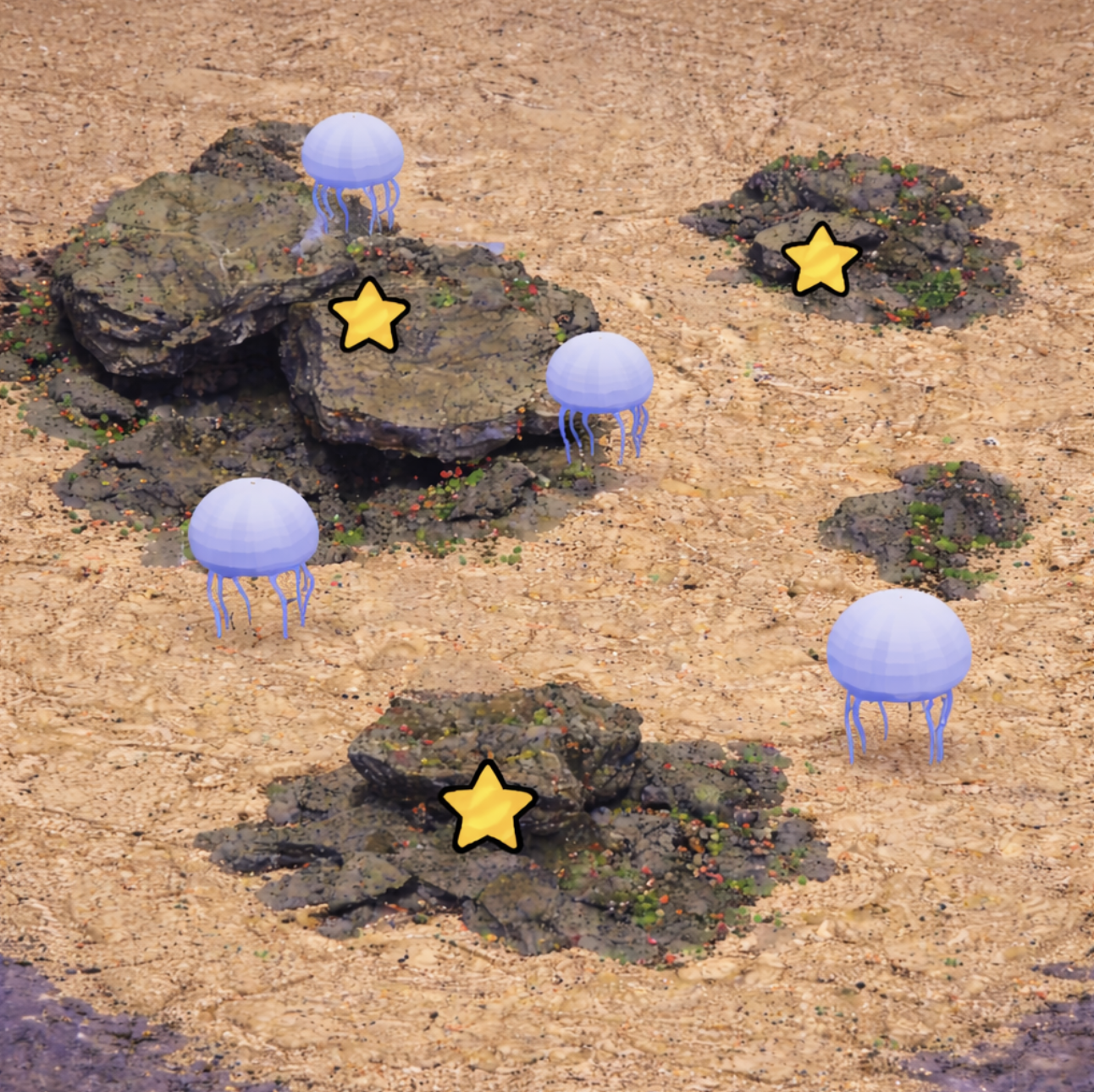}\\[-1mm]
  {\scriptsize (a) Salp}
\end{minipage}
&
\includegraphics[width=\linewidth]{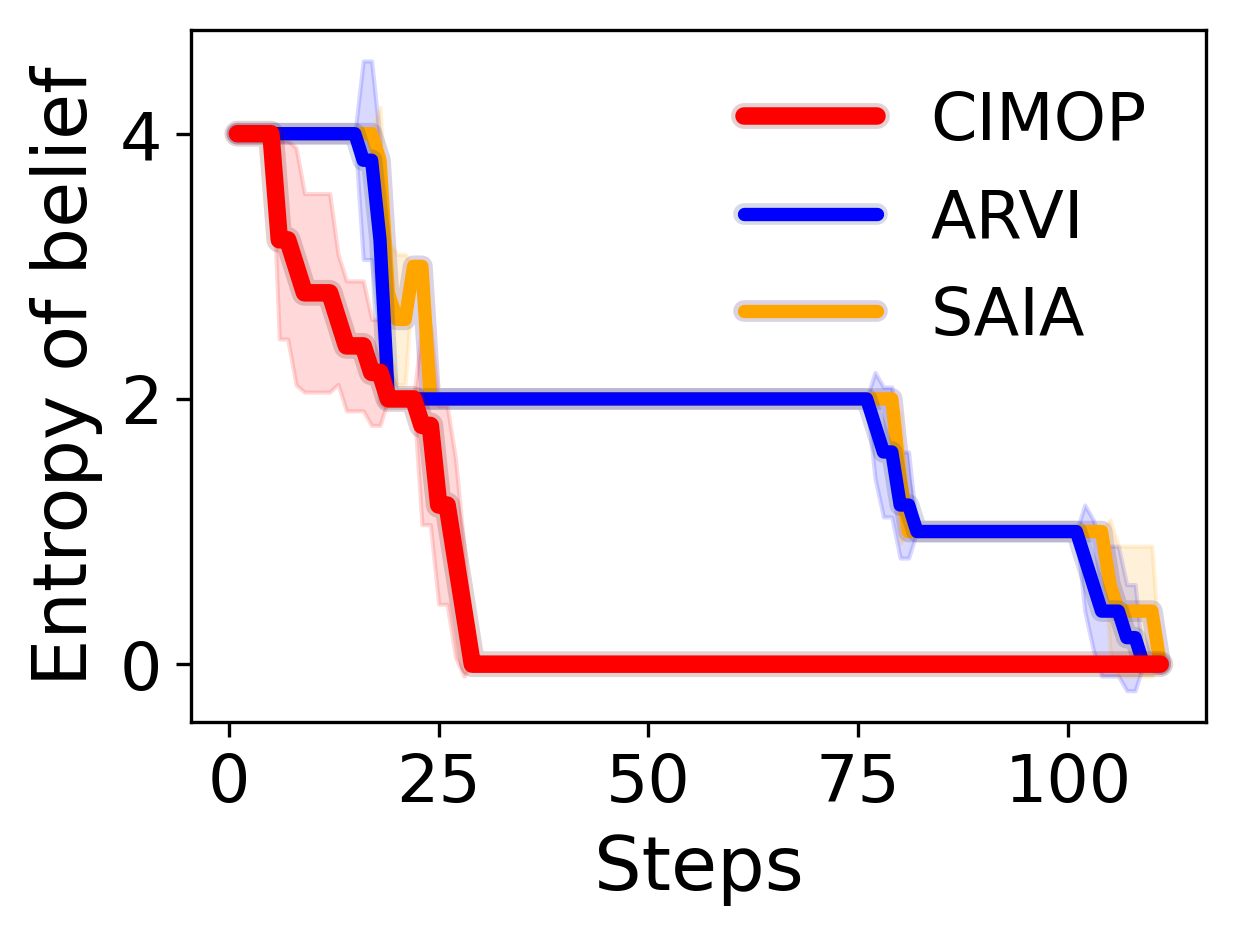}
&
\includegraphics[width=\linewidth]{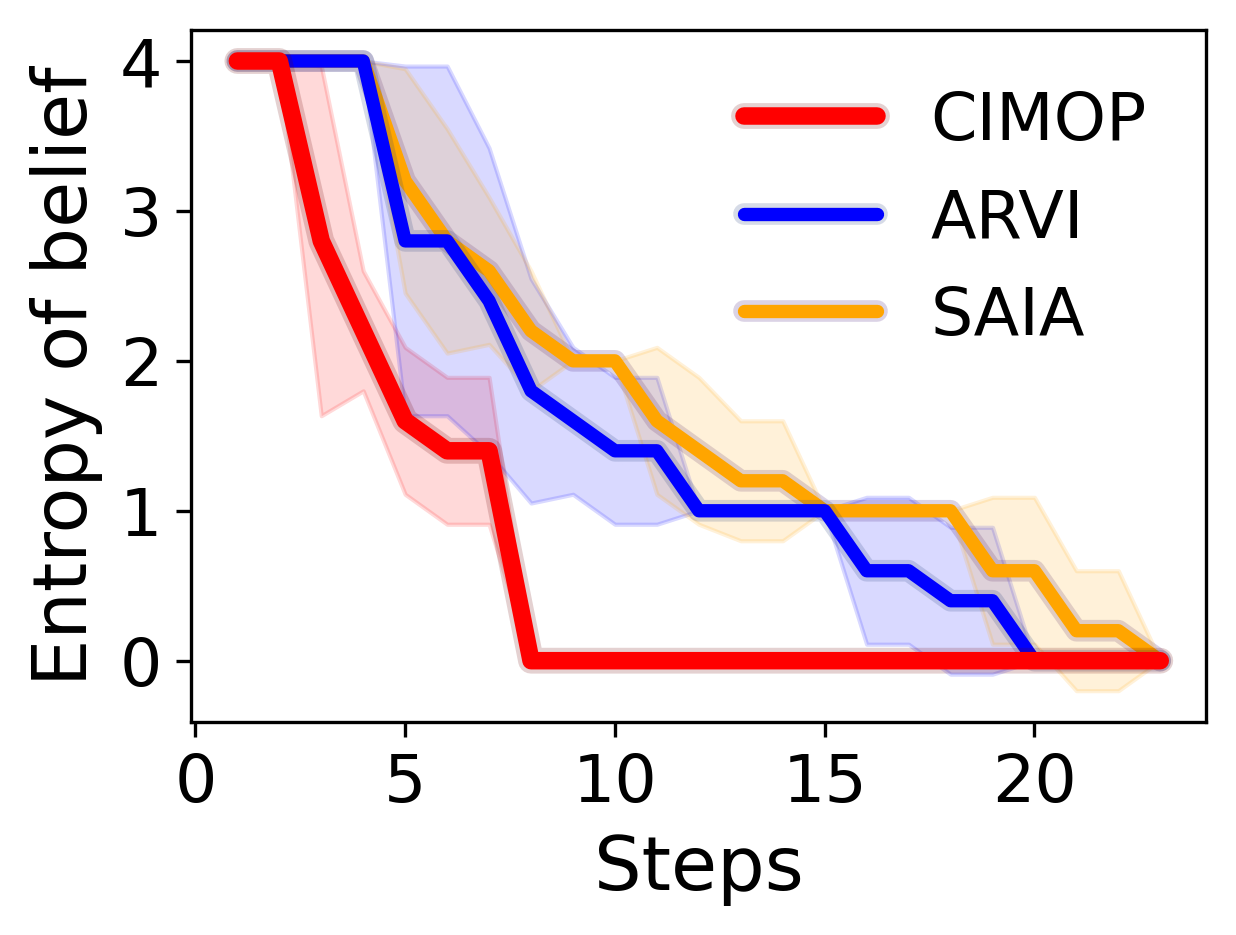}
&
\includegraphics[width=\linewidth]{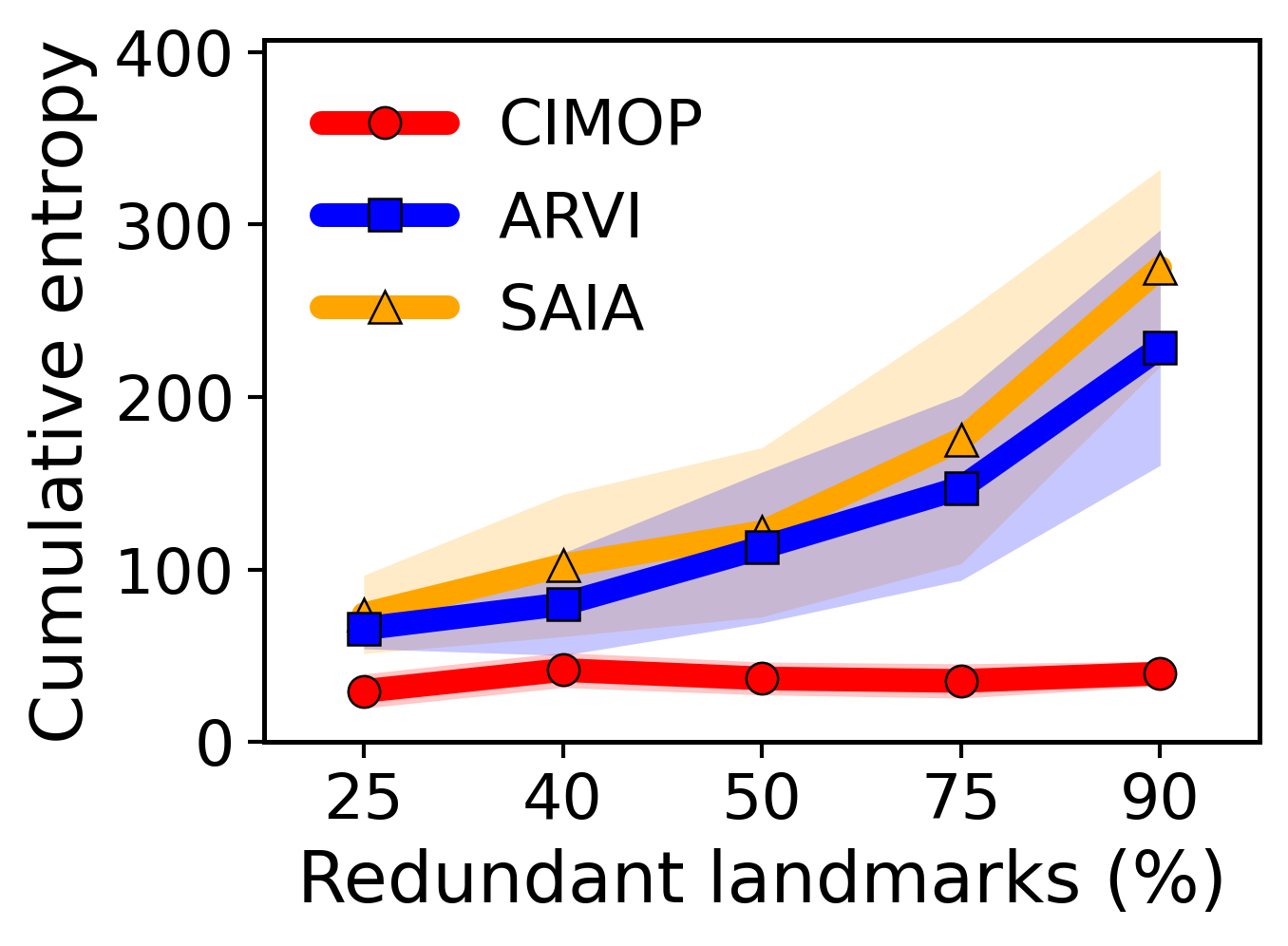}
&
\includegraphics[width=\linewidth]{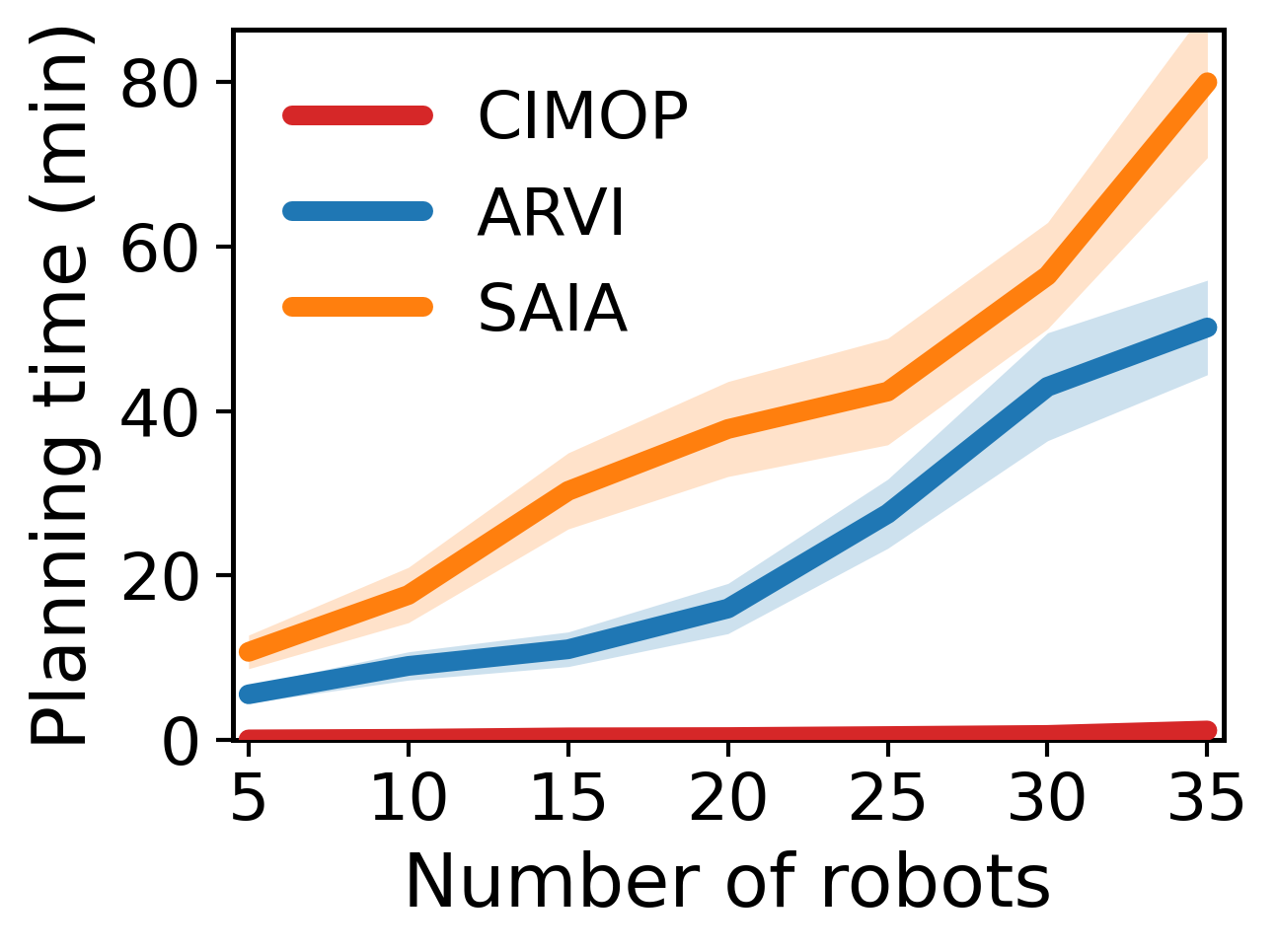}
\\

\begin{minipage}[c]{\linewidth}\centering
\vspace{-7mm}
  \includegraphics[width=0.8\linewidth]{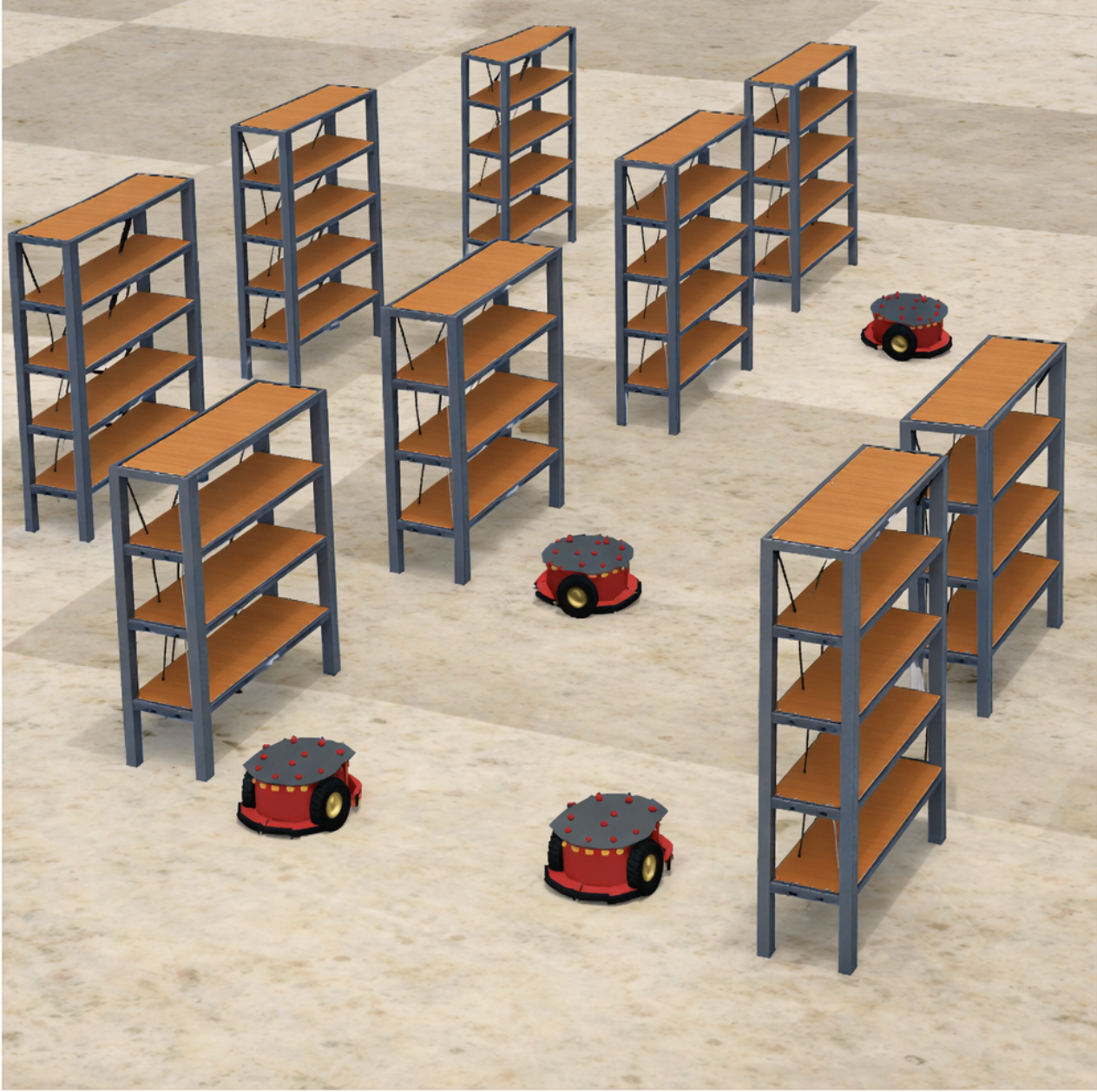}\\[-1mm]
  {\scriptsize (b) Warehouse}
\end{minipage}
&
\includegraphics[width=\linewidth]{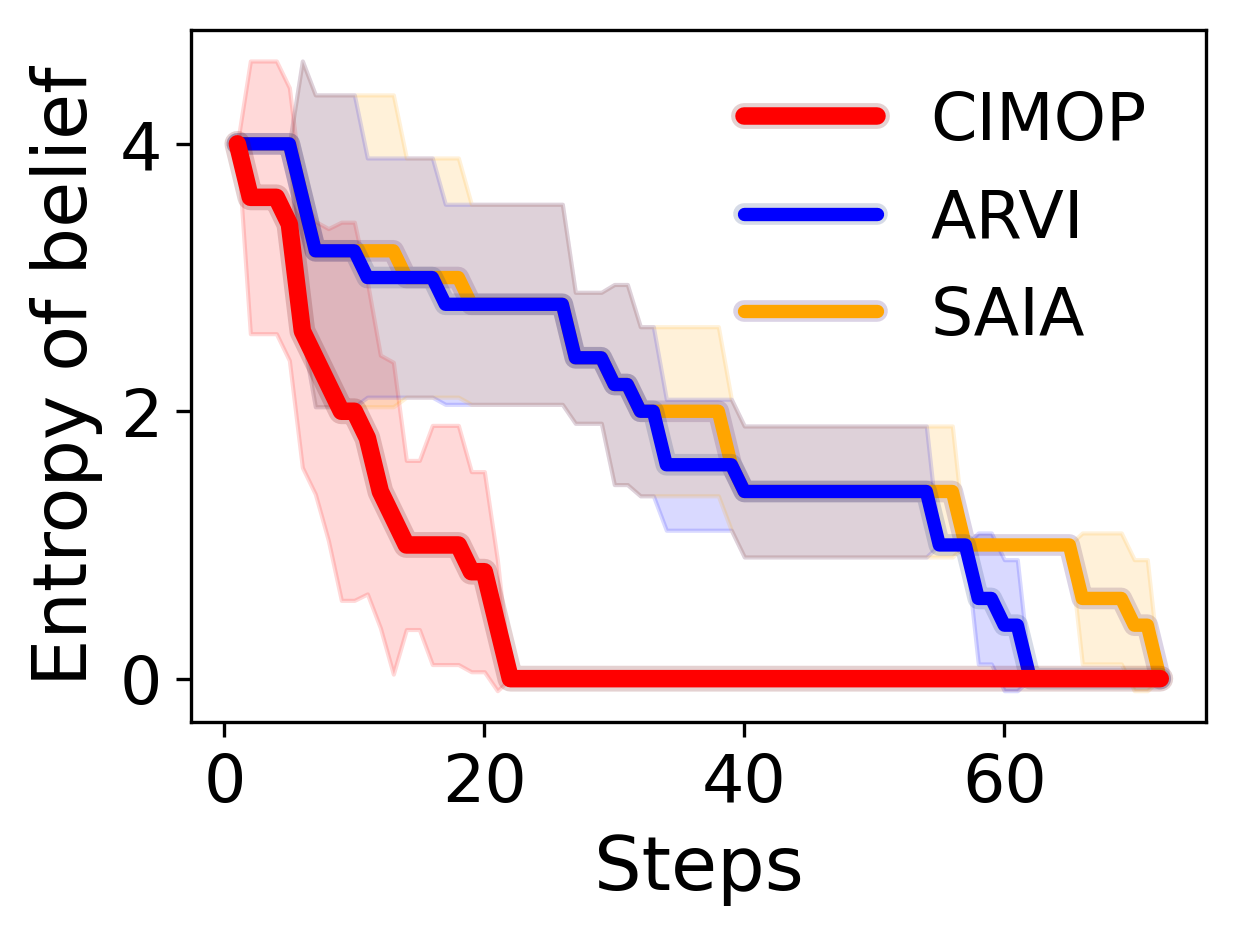}
&
\includegraphics[width=\linewidth]{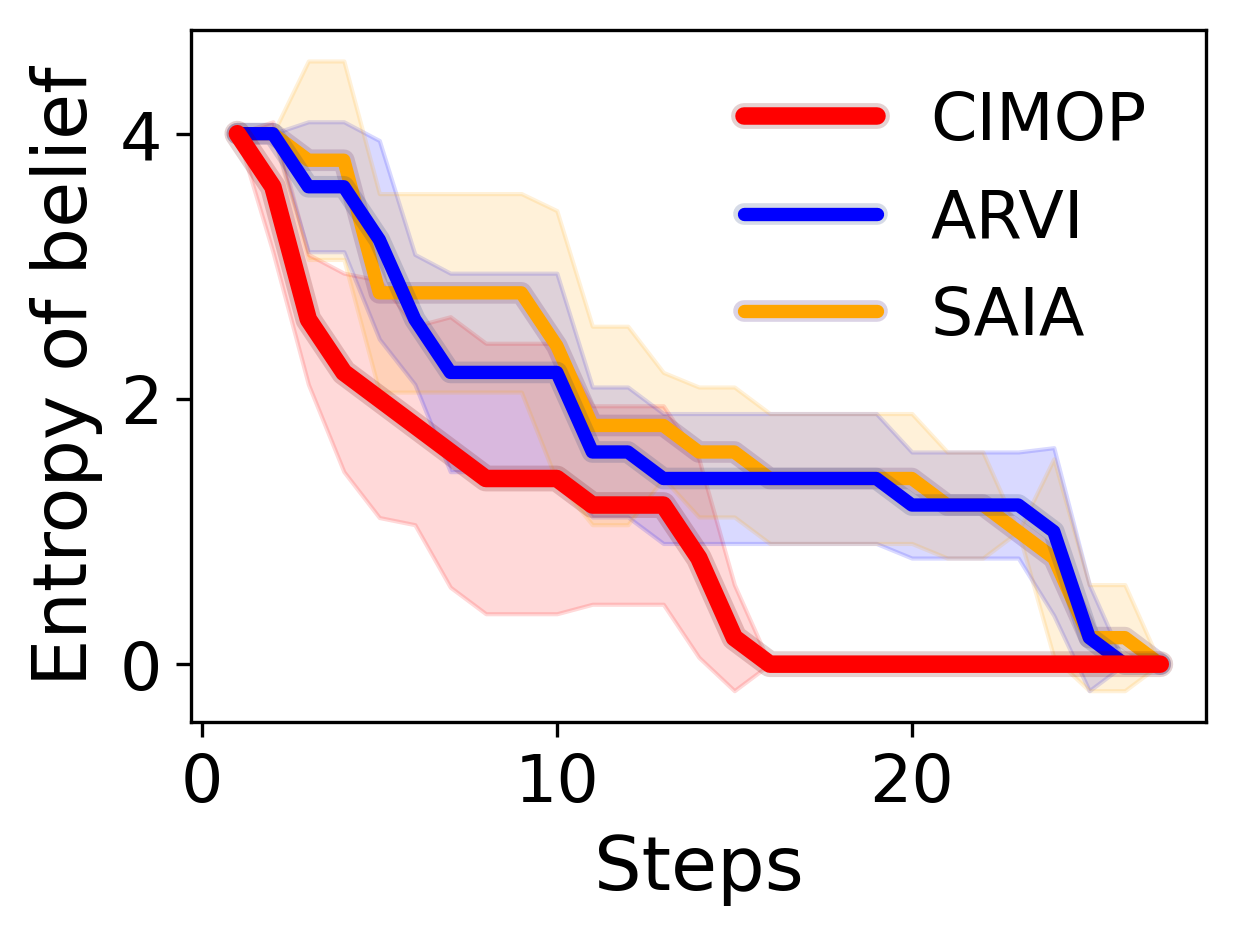}
&
\includegraphics[width=\linewidth]{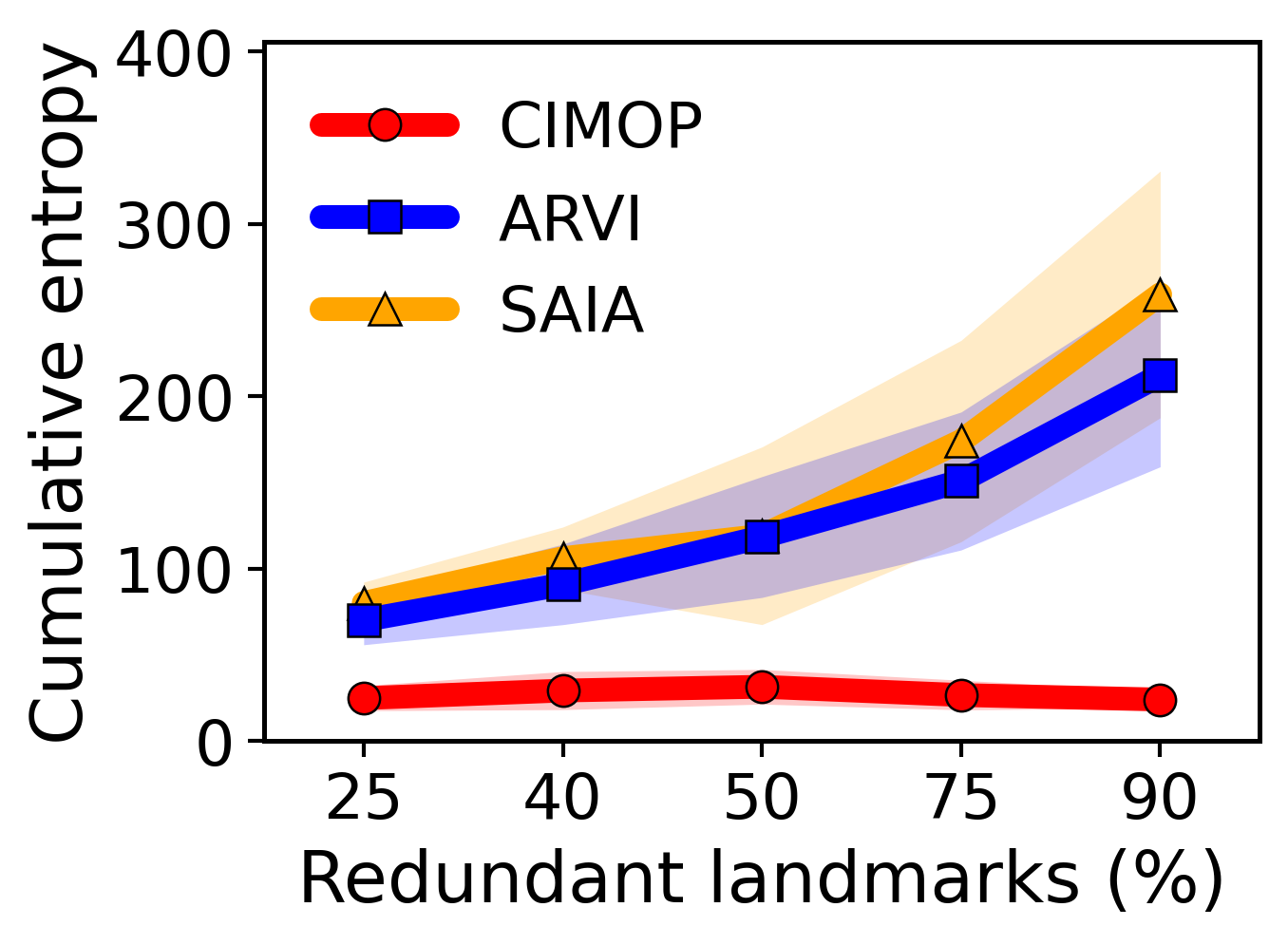}
&
\includegraphics[width=\linewidth]{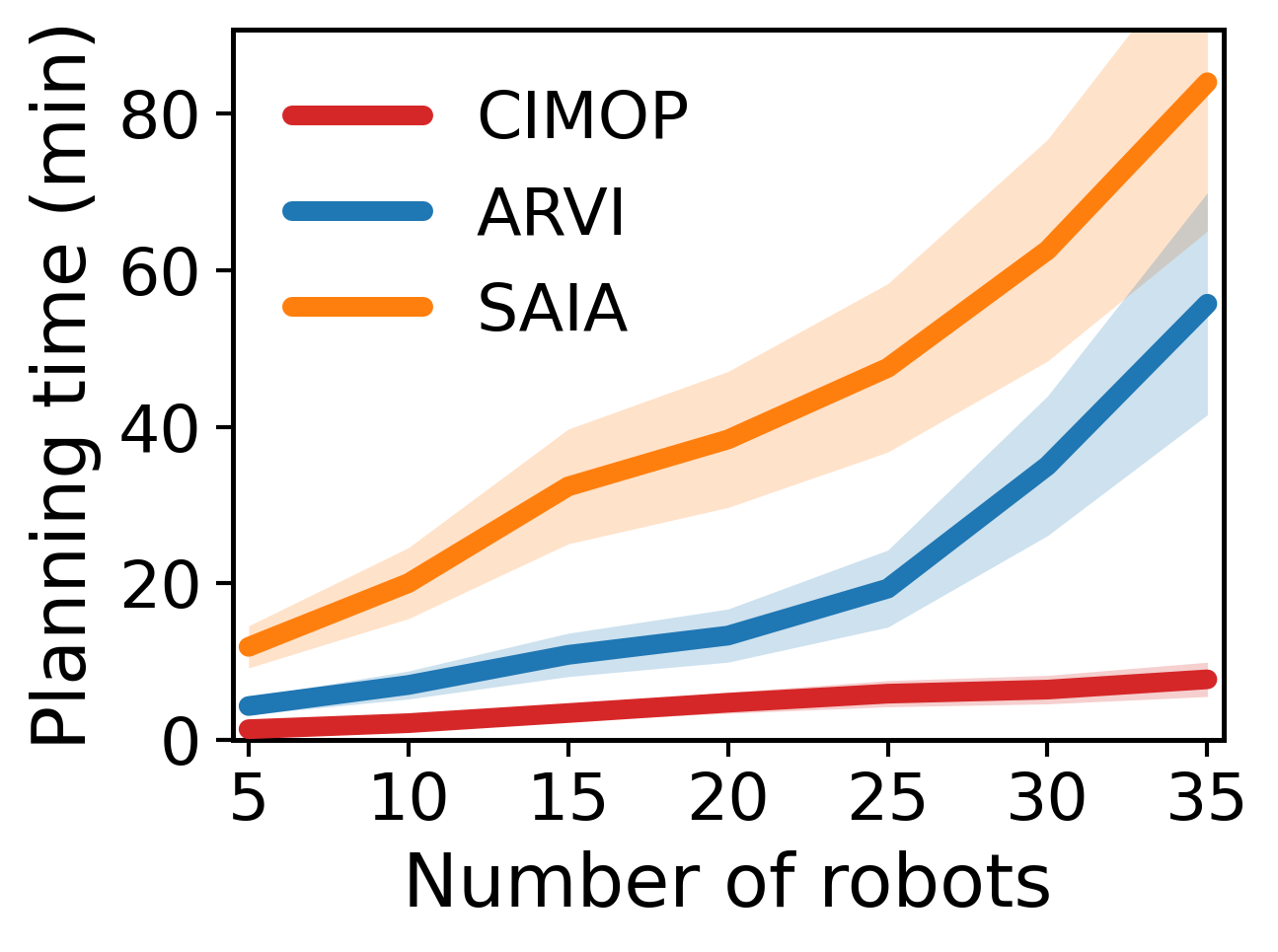}
\\

\begin{minipage}[c]{\linewidth}\centering
\vspace{-7mm}
  \includegraphics[width=0.8\linewidth]{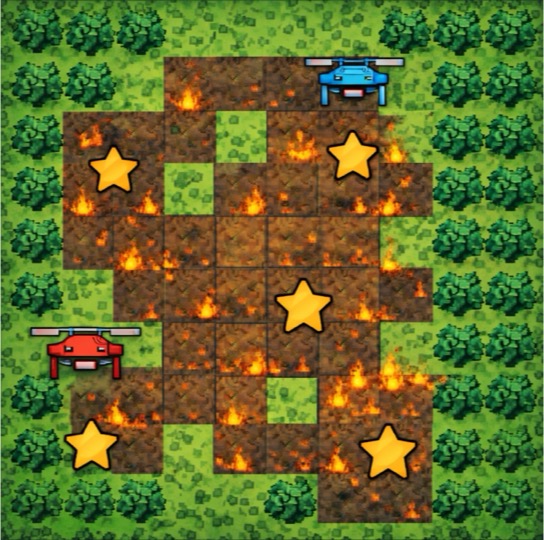}\\[-1mm]
  {\scriptsize (c) Forest firefighting}
\end{minipage}
&
\includegraphics[width=\linewidth]{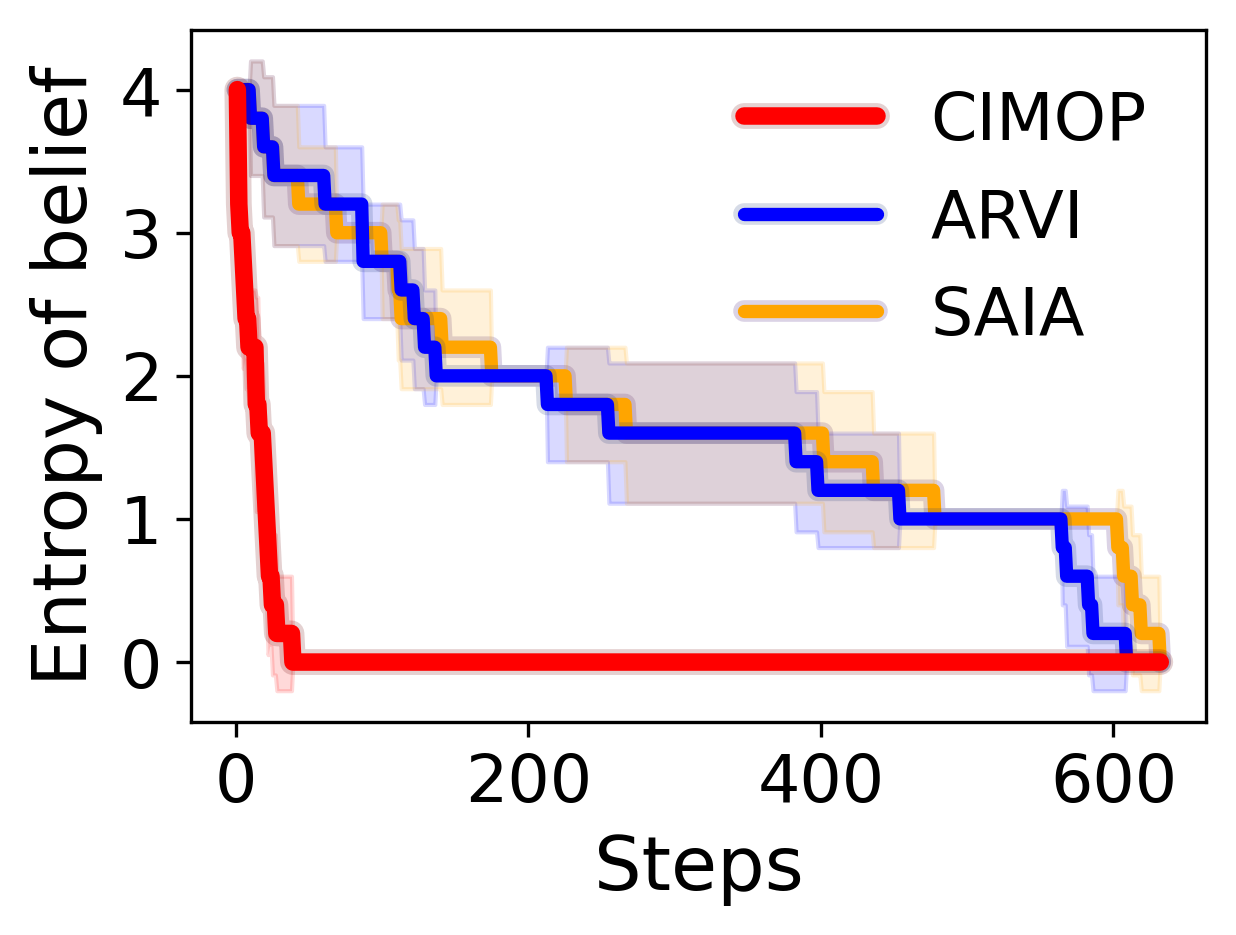}
&
\includegraphics[width=\linewidth]{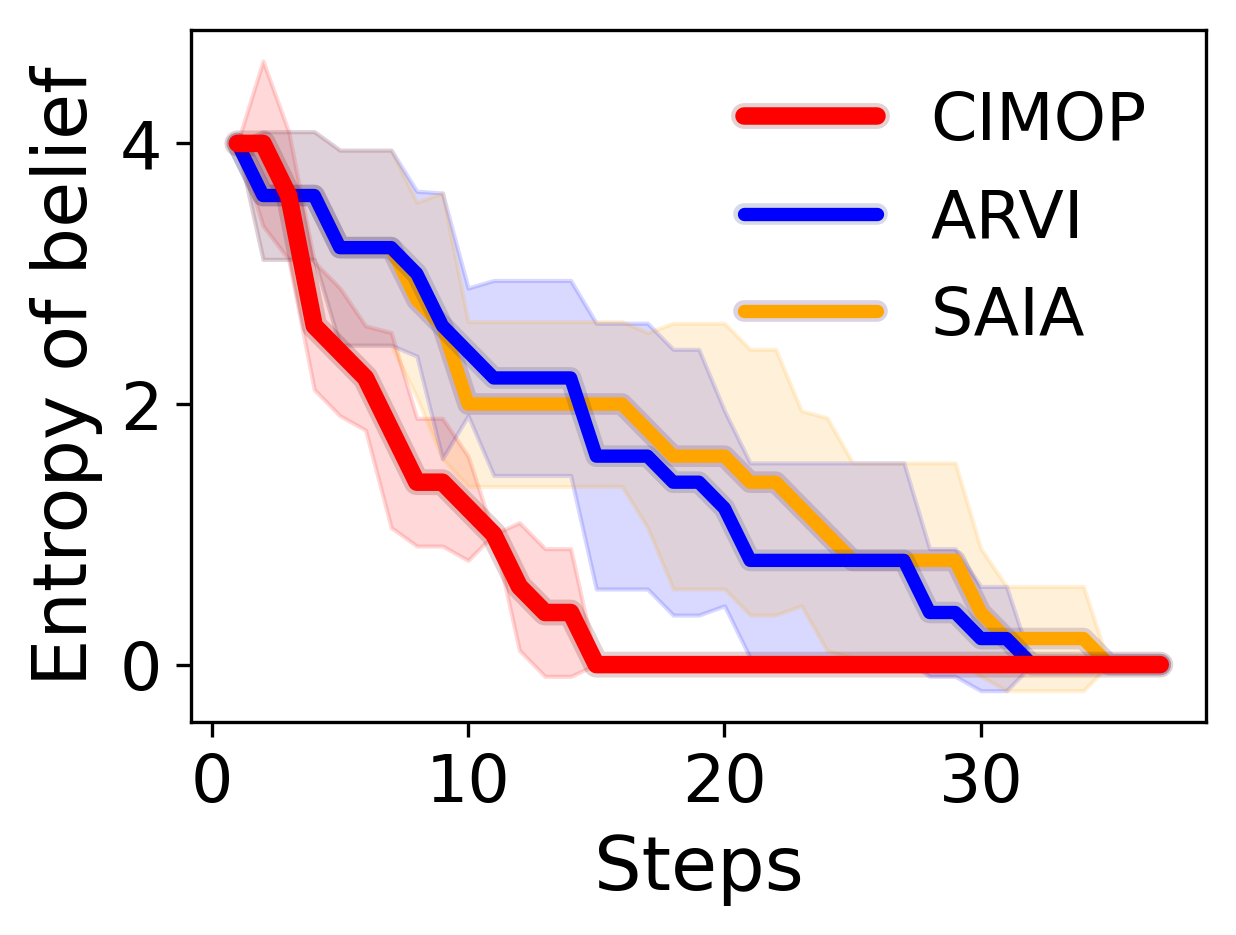}
&
\includegraphics[width=\linewidth]{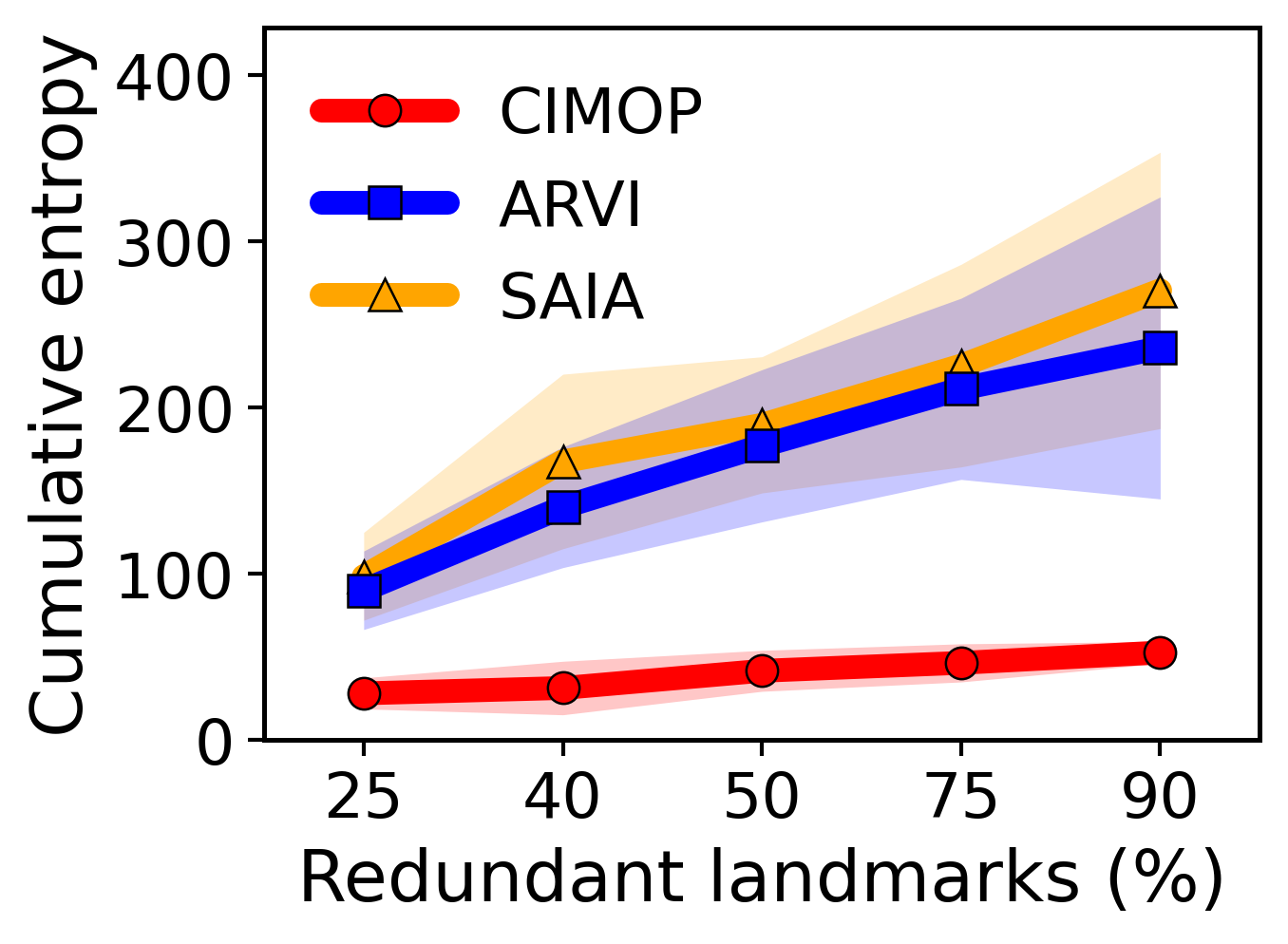}
&
\includegraphics[width=\linewidth]{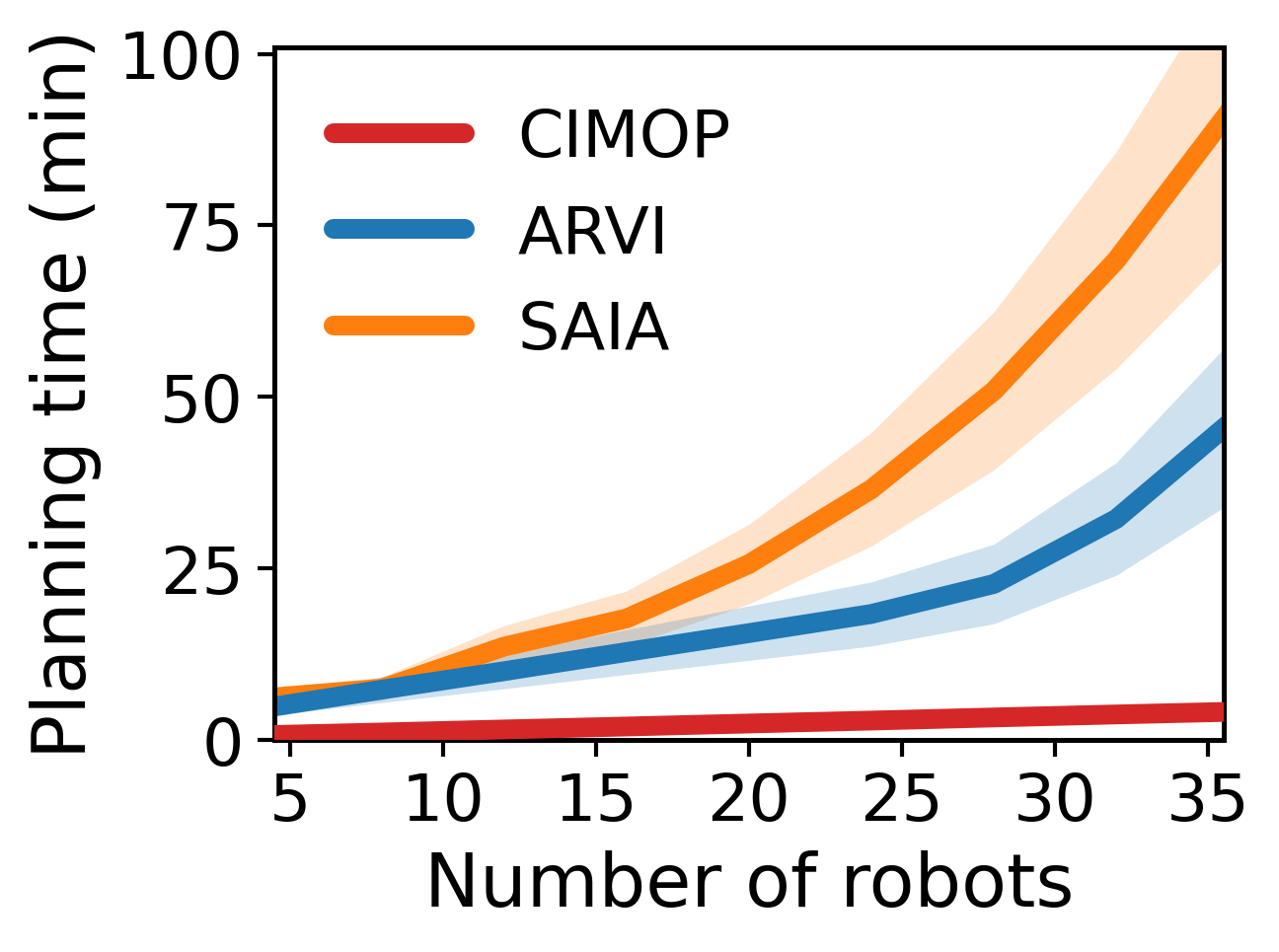}
\\

\bottomrule
\end{tabular}

\caption{Summary of experimental results across three domains. The first column shows the domain maps. The second and third columns show entropy-of-belief trends for $5$ and $35$ robots, respectively. The fourth column shows path cost with increasing percentage of redundant landmarks for analysis in cluttered environments. The fifth column shows plan computation time with increasing numbers of robots. All results are averaged over five instances with randomized robot starting locations.}

\label{fig:results_grid}
\end{figure*}
\section{EXPERIMENTS}
We evaluate our solution approach in simulation using three domains and on hardware using five mobile robots. Code: \url{https://tinyurl.com/MR-CUSSP}

\vspace{2pt}
\noindent \textbf{Baselines~} We compare the performance of our two-stage approach with combinations of state-of-the-art for each stage. Specifically, for the first stage, we compare CIMOP with (1) ARVI~\cite{schlotfeldt2018anytime} and (2) SAIA~\cite{kantaros2021sampling}. As both these approaches assume robots can gather information independently, we modify them to work with our observation function, for fair comparison. For the second stage, we compare LCBS against (1) BB-MO-CBS-$k\,(k=\!1)$~\cite{wang2024efficient} that returns a single non-dominated solution, serving as a fast Pareto baseline, (2) Scalarized CBS that uses scalarization in low level search using CBS~\cite{sharon2012conflict} to combine multiple objectives into a single weighted value, and (3) MO-CBS~\cite{ren2021multi} that maintains Pareto frontiers at the robot level and combines them through CBS-style conflict resolution, providing a principled Pareto-optimal multi-objective baseline without explicitly searching the full joint state space. 
For Scalarized CBS, scalarization weights follow geometric scaling: $C_1 M^{n-1} + C_2 M^{n-2} + \dots + C_n$, where $M$ must be sufficiently large to preserve $C_1 \succ \dots \succ C_n$. In practice, objective cost magnitudes are not known a priori, making weight selection non-trivial. Following~\cite{ho2023preference}, we sample $50$ random trajectories from start to goal to estimate the scale of each objective, and choose $M$ to exceed the largest observed lower-priority cost for a single instance in each domain. The same $M$ value is used for all other instances of that domain.

All algorithms were implemented in C++ and tested on a macOS machine with $18$ GB RAM. Robot evaluation is performed with five robots in the Robotarium testbed~\cite{wilson2020robotarium}. 

\noindent \textbf{Metrics~} We compare the algorithms using stage-specific metrics in simulation and the overall performance of the two-stage approach using mobile robots. The stage-specific metrics are: (i) Entropy of belief over contexts during execution (Eqn.~\ref{eq:entropy}); (ii) Cumulative entropy of the belief sequence to measure the duration of uncertainty until belief collapse (as opposed to uncertainty associated with a particular belief as in (i)); (iii) Scalability with increasing number of robots; and (iv) Performance with varying planning time budget.

\subsection{Evaluation in simulation}
In the following domains, robots begin with a uniform initial belief over a finite set of contexts and coordinate to obtain informative observations at landmark states.

\noindent\emph{a) Sample Collection with Salps:} In this domain, salp-inspired~\cite{sutherland2010comparative} underwater robots are tasked with collecting samples in environments with crevices and boulder ridges that act as informative landmarks. Robots begin with a uniform belief over three contexts: $c_1$ (strong-current region), $c_2$ (coral-sensitive region), and $c_3$ (nominal condition). In all contexts, robots should minimize energy consumption ($o_e$), minimize coral damage ($o_c$), and minimize time to goal ($o_t$), but their priority depends on the context. The context to objective ordering is $c_1\!:\!o_e\!\succ\!o_c\!\succ\!o_t$; $c_2\!:\!o_c \succ o_e \succ o_t$; and $c_3\!:\!o_t \succ o_e \succ o_c$. Coordinated sampling at landmarks reveals flow and surface characteristics that distinguish among these contexts. Across instances, we vary the number and spatial distribution of informative landmarks and robots.

\noindent\emph{b) Heavy-Lift Warehouse:}
Autonomous warehouse robots transport shelves through shared aisles to fulfillment stations, with designated monitoring stations serving as informative landmarks. Robots begin with a uniform belief over three contexts: $c_1$ (backlogged), $c_2$ (congested), and $c_3$ (human-traffic). In all contexts, robots balance completion time ($o_t$), minimizing congestion in primary aisles, ($o_c$) and avoiding human zones ($o_h$), but the priority depends on context. The context to objective ordering is as follows: $c_1\!:\!o_t \succ o_c \succ o_h$; $c_2\!:\!o_c\succ o_t\succ o_h$; and $c_3\!:\!o_h\succ o_c\succ o_t$. At monitoring stations, two robots must arrive simultaneously at paired checkpoints to obtain a joint observation of aisle-level traffic and human presence, which cannot be determined from a single robot's local view. Across instances, we vary the number of robots, shelf locations, and the placement of primary aisles and monitoring stations.

\noindent\emph{c) Forest Firefighting:} 
Aerial robots are assigned to reach designated fire-control locations in an evolving wildfire. Since large-scale wind and smoke conditions cannot be determined from local sensing alone, robots coordinate at designated vantage points that serve as informative landmarks before planning traversal. Robots begin with a uniform belief over three contexts: $c_1$ (high-wind), $c_2$ (dense-smoke), and $c_3$ (rapid-spread). In all contexts, robots should minimize completion time ($o_t$), minimize energy consumption ($o_e$), and minimize path length ($o_l$), but the priority depends on the context. The context to objective ordering is $c_1\!:\!o_e \succ o_l \succ o_t$; $c_2\!:\!o_l \succ o_e \succ o_t$; and $c_3\!:\!o_t \succ o_e \succ o_l$. Coordinated observations at vantage points reveal wind and smoke conditions that distinguish among these contexts. Across instances, we vary the number of robots, vantage point placement, and the locations of wind and fire spots.

\begin{figure}[t]
  \centering
  \includegraphics[width=0.84\linewidth]{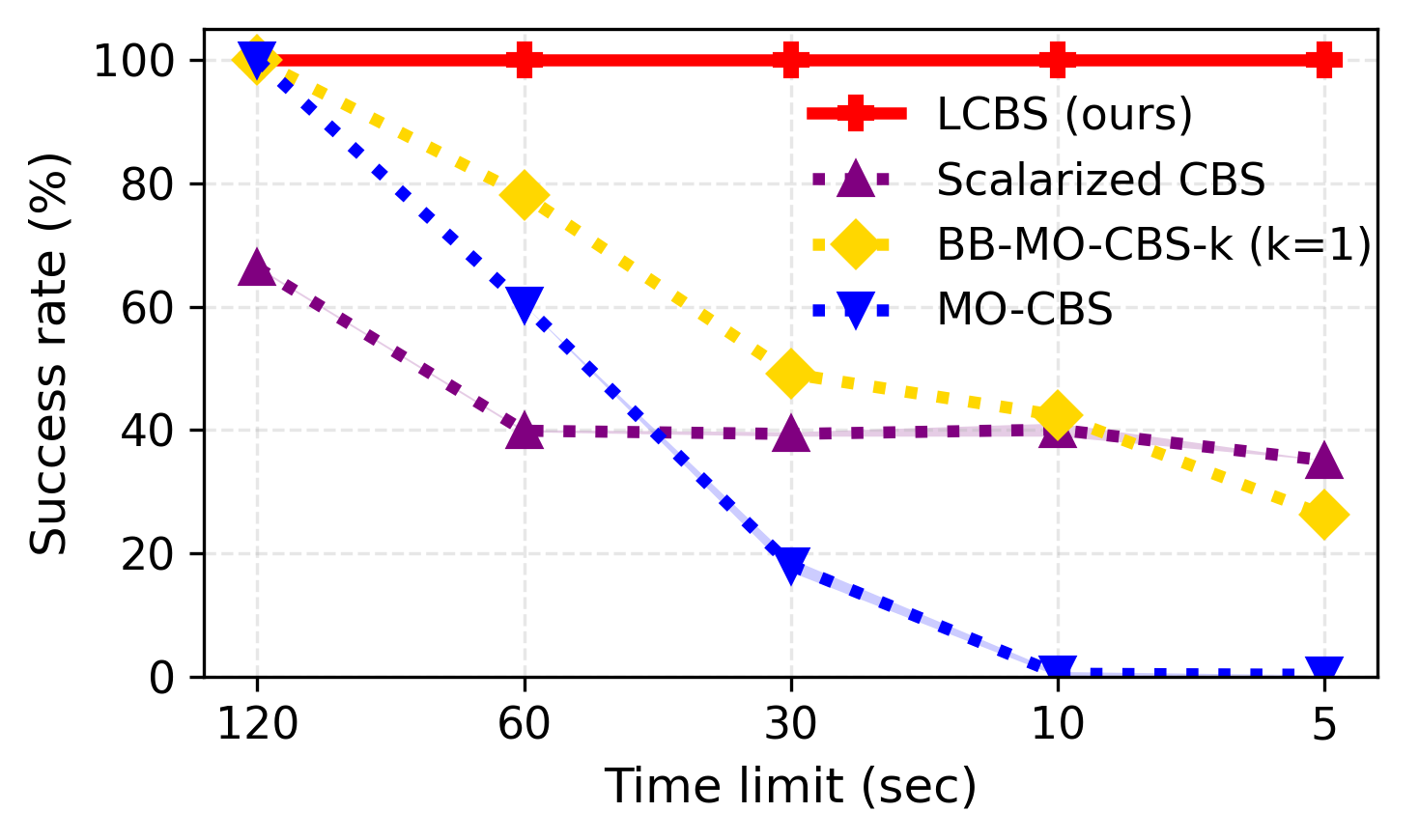}
  \caption{Success rate on 15 instances (five from each of the three domains) under constrained planning time limit with five robots and three objectives.}
  \label{fig:optimality-lines}
\end{figure}

\section{RESULTS and DISCUSSION}

\noindent\emph{a) Belief entropy reduction: }
Fig.~\ref{fig:results_grid} (columns (i)–(ii)) reports belief entropy during execution, computed using Eqn.~\ref{eq:entropy}, where lower values indicate faster context inference. Across all settings, CIMOP consistently infers the context faster than both baselines. For five robots, CIMOP reaches zero entropy within 20 steps in the salp and warehouse domains, compared to 60–80 steps for the baselines, and collapses substantially faster in the forest firefighting domain where baselines require several hundred steps. While we observe an improved performance with $35$ robots, across methods, CIMOP remains consistently faster in context inference.

\vspace{2pt}
\noindent\emph{b) Cumulative entropy with increasing redundant landmarks: }
To quantify how long it takes to infer the true context and whether useful landmarks are prioritized correctly over time, we measure the cumulative entropy of a belief sequence induced by the policy. This complements the entropy of a single belief point (Eqn.~\ref{eq:entropy}) and evaluates the sequential nature of uncertainty reduction. We measure this by varying the percentage of landmark states whose associated observations are \emph{redundant} with respect to observations available at other landmark states and therefore visiting them does not reduce belief entropy. Figure~\ref{fig:results_grid} (column (iii)) shows cumulative entropy, averaged over five instances in each domain, with increasing percentage of redundant landmarks. Lower cumulative entropy indicates faster inference of the true context. CIMOP maintains consistently low cumulative entropy across all domains and clearly outperforms baselines, with low variance despite sparse informative observations.

\vspace{2pt}
\noindent\emph{c) Planning time: }
Figure~\ref{fig:results_grid} (column (iv)) evaluates scalability through wall-clock planning time for context inference. CIMOP scales substantially better than both baselines in all domains. For $35$ robots, CIMOP consistently plans in under two minutes, while ARVI exceeds $40$ minutes and SAIA exceeds $80$ minutes. This consistent gap demonstrates CIMOP's ability to scale to larger robot teams while maintaining practical planning times.

\begin{figure}[t]
    \centering 
    \includegraphics[width=0.95\linewidth]{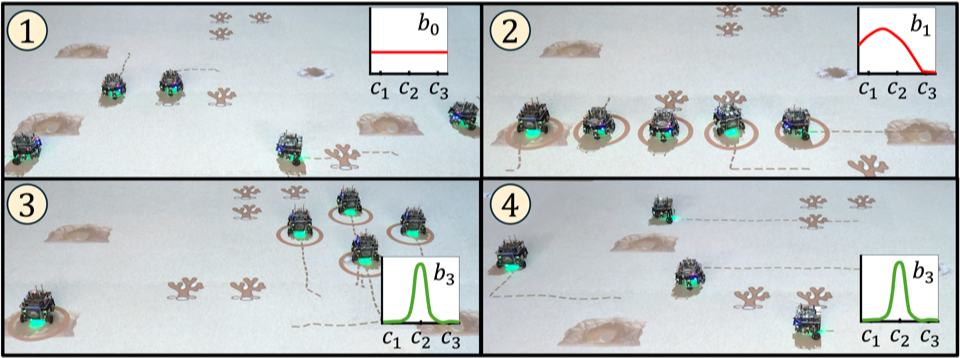}
    \caption{Execution sequence with five GTernal robots~\cite{wilson2020robotarium}: (1) start with uniform belief over three contexts, (2) form a chain to observe the cave landmark and update belief, (3) four robots form a ring to observe the crevice landmark and collapse belief, after which (4) the robots independently plan prioritizing minimizing coral damage over energy consumption and time to goal as imposed by the inferred context $c_2$.}
    \label{fig:salp_setup}
\end{figure}

\begin{figure}[!t]
    \centering
    \includegraphics[width=\linewidth]{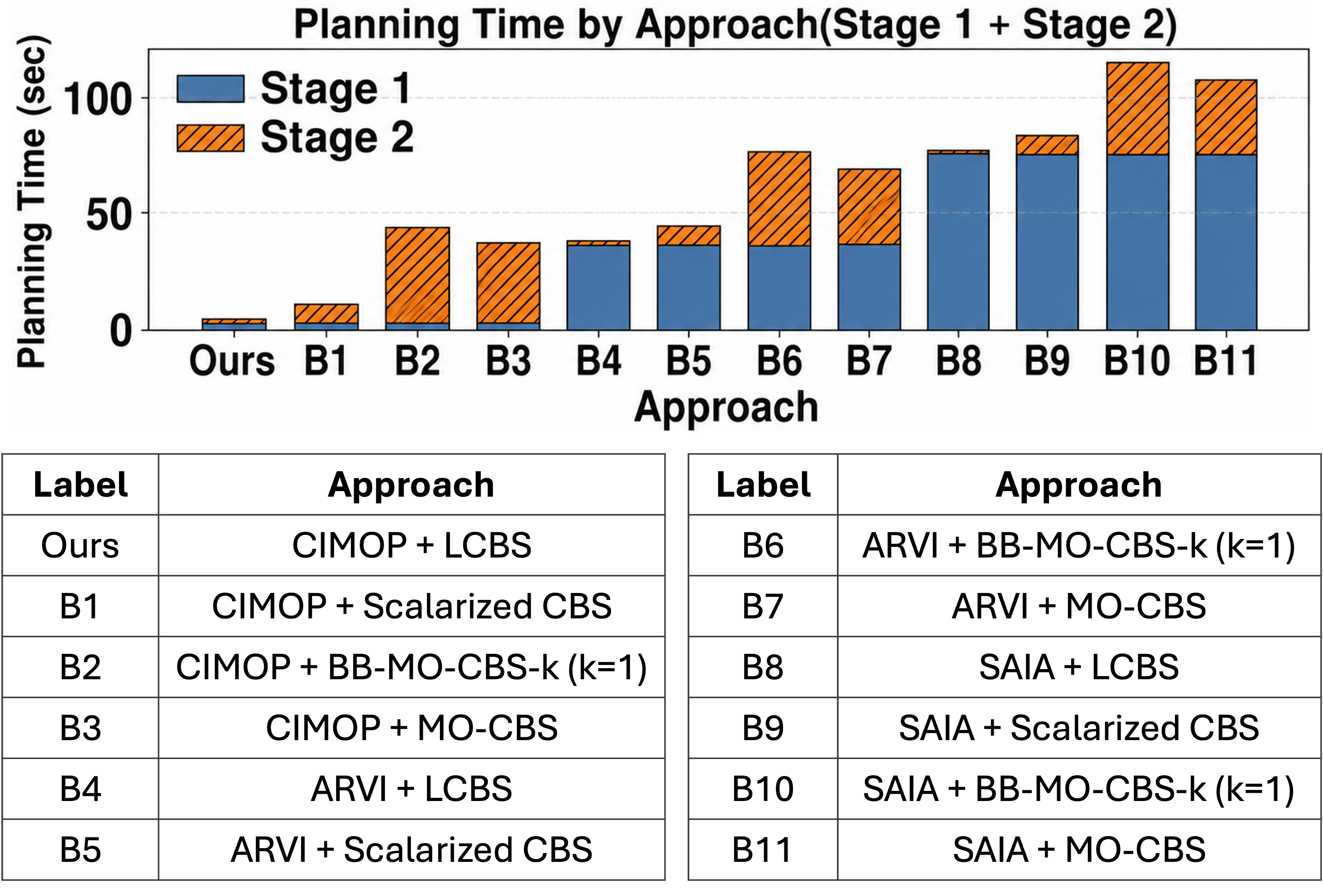}
    \caption{Total time taken to plan by each approach in our hardware experiment with five mobile robots and three possible contexts. Each context is characterized by a different preference ordering over three objectives.}
    \label{fig:total_planning_time}
\end{figure}

\vspace{2pt}
\noindent\emph{d) Performance in time-constrained settings: }
The efficiency of planning in time-constrained settings is measured in terms of \emph{success rate}~\cite{wang2024efficient,ren2021multi}: the fraction of problem instances where the algorithm was able to generate an optimal solution for the lexicographic ordering over objectives, which can be verified by comparing the results with a Pareto-frontier approach (such as MO-CBS).  Fig.~\ref{fig:optimality-lines} reports the success rate on 15 environment instances (five from each of the three domains) as the planning-time budget decreases from $120$ to $5$ seconds for five robots and three objectives. LCBS maintains $100\%$ success across all budgets and remains consistent even at five seconds, while BB-MO-CBS ($k\!=\!1$), MO-CBS, and Scalarized CBS drop significantly as the time limit decreases. The consistent performance of LCBS under tight budgets highlights that eliminating lexicographically dominated nodes during search allows it to compute preference-aligned solutions reliably in time-sensitive settings.

\noindent\emph{e) Hardware experiments: }
We validate the practical feasibility of our approach using five mobile robots in the salp domain. Fig.~\ref{fig:salp_setup} shows shared belief evolution, where robots adapt formations based on observation requirements at landmarks, and collapse the belief in real time before executing preference-aligned plans. Fig.~\ref{fig:total_planning_time} reports total planning time (stage 1 + stage 2), where CIMOP + LCBS completes both stages in under $6$ seconds, while other combinations require $40$–$120$ seconds. Combinations using LCBS maintain low second-stage times, and CIMOP substantially reduces context inference time, thus demonstrating the run time benefits of both CIMOP and LCBS which enable real-time execution on physical robots.

\section{SUMMARY AND FUTURE WORK}

This paper formalizes multi-robot planning under a fixed but initially unknown context as an MR-CUSSP. We present two algorithms that enable robot coordination to obtain informative joint observations to infer the operative context, and then compute collision-free plans aligned with the induced lexicographic preference ordering over objectives. 
Experimental results across three simulated domains and hardware deployments demonstrate faster belief collapse, lower cumulative entropy, and improved scalability in planning time with increasing robots and objectives compared to state-of-the-art baselines for each stage in our solution approach. The current formulation assumes that landmarks are predefined and observations at landmarks are either belief-collapsing or non-informative. In the future, we will relax these assumptions and extend our framework to support noisy observations at landmarks, and investigate how this may impact task performance.

\section*{ACKNOWLEDGMENTS}
This work was supported in part by ONR grant N00014-23-1-217, NSF grant 2543646, and DARPA TIAMAT grant HR0011-24-9-0423.

\bibliographystyle{IEEEtran.bst}
\bibliography{references}
\end{document}